\def\paperlanguage{}
\pdfoutput=1 % for arxiv

\documentclass[letterpaper, 10 pt, conference]{ieeeconf}  % Comment this line out if you need a4paper

\usepackage{bm}
\usepackage{cite}
\usepackage{flushend}
\include{preamble}

\IEEEoverridecommandlockouts                              % This command is only needed if
\title{\LARGE \textbf
  {
    \switchlanguage%
    {%
      Component Modularized Design of Musculoskeletal Humanoid Platform Musashi to Investigate Learning Control Systems
    }%
    {%
      学習型制御模索のための筋骨格ヒューマノイドプラットフォームMusashiのモジュラー型設計
    }%
  }
}

\author{Kento Kawaharazuka$^{1}$, Shogo Makino$^{1}$, Kei Tsuzuki$^{1}$, Moritaka Onitsuka$^{1}$, Yuya Nagamatsu$^{1}$\\Koki Shinjo$^{1}$, Tasuku Makabe$^{1}$, Yuki Asano$^{1}$, Kei Okada$^{1}$, Koji Kawasaki$^{2}$ and Masayuki Inaba$^{1}$% <-this % stops a space
  \thanks{$^{1}$ The authors are with Department of Mechano-Informatics, Graduate School of Information Science and Technology, The University of Tokyo, 7-3-1 Hongo, Bunkyo-ku, Tokyo, 113-8656, Japan.
    {\texttt\small [kawaharazuka, makino, tsuzuki, onitsuka, nagamatsu, shinjo, makabe, asano, k-okada, inaba]@jsk.t.u-tokyo.ac.jp}
  }
  \thanks{$^{2}$ The author is associated with TOYOTA MOTOR CORPORATION.
    {\texttt\small koji\_kawasaki@mail.toyota.co.jp}
  }
}
\begin{document}

\maketitle
\thispagestyle{empty}
\pagestyle{empty}

%%%%%%%%%%%%%%%%%%%%%%%%%%%%%%%%%%%%%%%%%%%%%%%%%%%%%%%%%%%%%%%%%%%%%%%%%%%%%%%%
\begin{abstract}
  \switchlanguage%
  {%
    To develop Musashi as a musculoskeletal humanoid platform to investigate learning control systems, we aimed for a body with flexible musculoskeletal structure, redundant sensors, and easily reconfigurable structure.
    For this purpose, we develop joint modules that can directly measure joint angles,  muscle modules that can realize various muscle routes, and nonlinear elastic units with soft structures, etc.
    Next, we develop MusashiLarm, a musculoskeletal platform composed of only joint modules, muscle modules, generic bone frames, muscle wire units, and a few attachments.
    Finally, we develop Musashi, a musculoskeletal humanoid platform which extends MusashiLarm to the whole body design, and conduct several basic experiments and learning control experiments to verify the effectiveness of its concept.
  }%
  {%
    本研究では, 学習制御を模索するためのハードウェアプラットフォームMusashiの要件として, 柔軟な身体を持つ筋骨格系, 冗長なセンサ, 容易に構造を変化させることのできる身体を目指した.
    そこで, 関節状態を測定できる関節モジュール, 汎用的に筋経路を実現可能な筋モジュール, それ自体が柔軟な非線形弾性ユニット等を開発する.
    次に, 関節モジュール・筋モジュール・汎用骨格・筋ワイヤユニット・少数のアタッチメントのみで簡易に構成可能な筋骨格プラットフォームMusashiLarmを開発した.
    最後に, MusashiLarmを全身に拡張した筋骨格ヒューマノイドプラットフォームMusashiの設計開発について述べ, いくつかの予備実験と学習制御を行い, その有用性を確認する.
  }%
\end{abstract}

%%%%%%%%%%%%%%%%%%%%%%%%%%%%%%%%%%%%%%%%%%%%%%%%%%%%%%%%%%%%%%%%%%%%%%%%%%%%%%%%
\section{INTRODUCTION} \label{sec:1}
\begin{figure}[t]
  \centering
  \includegraphics[width=1.0\columnwidth]{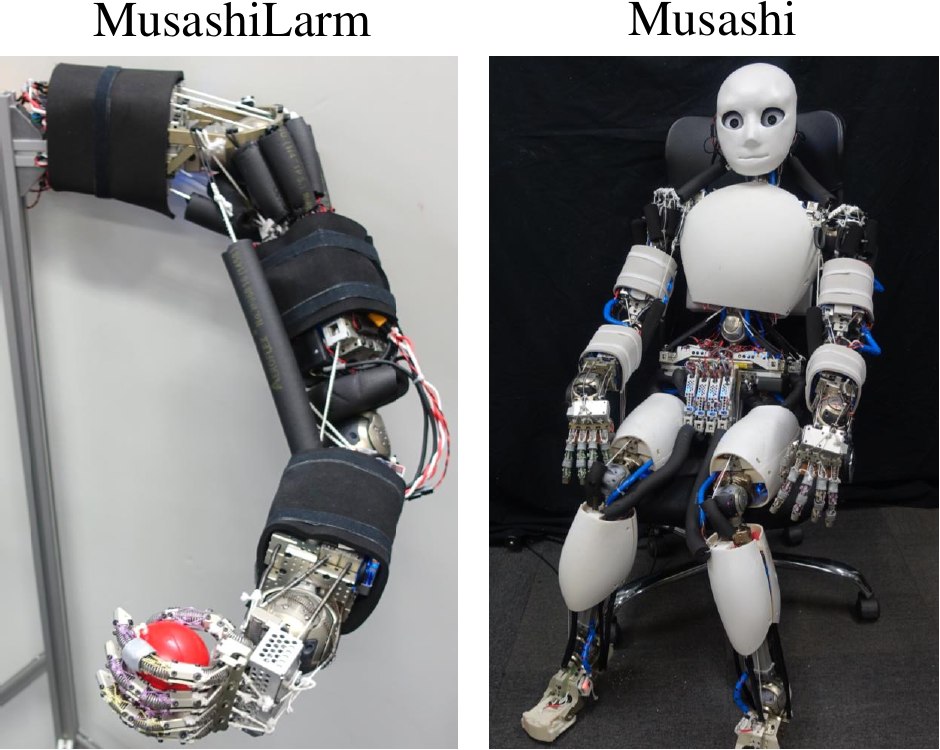}
  \vspace{-3.0ex}
  \caption{The newly developed MusashiLarm and Musashi.}
  \label{figure:design-all}
  \vspace{-1.0ex}
\end{figure}

\switchlanguage%
{%
  The tendon-driven musculoskeletal humanoid \cite{nakanishi2013design, wittmeier2013toward, jantsch2013anthrob, asano2016kengoro}, which imitates not only the human proportion but also the joint and muscle structure of human beings, has many benefits.
  For example, it can realize mechanical variable stiffness, measure muscle tensions by using not expensive torque sensors but pressure sensors or loadcells, apply ball joints that have no singularity, and use an underactuated system such as the spine.
  Also, its body, which has a muscle structure like the human body, is useful in the sense that we can understand the human body better, embed humanlike learning control systems, and realize more human-like motions.

  In this study, we develop Musashi as a hardware platform to investigate learning control systems of musculoskeletal structures such as self-body image acquisition methods \cite{kawaharazuka2018online, kawaharazuka2018bodyimage} (\figref{figure:design-all}).
  We consider 3 requirements for such a musculoskeletal humanoid platform.
  \begin{itemize}
    \item It is a musculoskeletal humanoid with a flexible body structure.
    \item It has redundant sensors to investigate learning control systems.
    \item Its construction and reconstruction are easy, and we can transform the body structure.
  \end{itemize}
  To develop a musculoskeletal humanoid platform with such characteristics has 3 benefits.
  \begin{itemize}
    \item We can safely investigate learning control systems using the flexible body and variable stiffness control.
    \item Although joint encoders usually cannot be installed in the musculoskeletal humanoid, a mechanism to directly measure joint angles makes experimental evaluations easy and accelerates developments.
    \item By changing link lengths, increasing degrees of freedom (DoFs), adding elastic elements, etc., we can consider changes in self-body image due to growth and deterioration of the body components.
  \end{itemize}
}%
{%
  人体の筋骨格構造をを模倣した筋骨格ヒューマノイド\cite{nakanishi2013design, wittmeier2013toward, jantsch2013anthrob, asano2016kengoro}は, 冗長な筋肉を用いてハードウェアで可変剛性を実現できる点, 高価なトルクセンサでなく圧力センサやロードセル等によって筋長力を計測できる点, 特異点のない球関節を使用できる点, 柔軟な背骨等の劣駆動系を採用できる点など, 多くの優れた点を持つ.
  また, 人間のように筋肉によって構成された体は, 人間を知る, 人間型の学習制御等を考える, という意味でも有用であると考える.

  本研究では, 自己身体像獲得\cite{kawaharazuka2018online, kawaharazuka2018bodyimage}等の筋骨格ヒューマノイドにおける学習制御を模索するためのプラットフォームとして, Musashiを開発する(\figref{figure:design-all}).
  筋骨格ヒューマノイドプラットフォームの条件として以下の3つを考えた.
  \begin{itemize}
    \item 柔軟な身体構造を有する筋骨格ヒューマノイドである.
    \item センサが冗長に存在し, 学習制御模索のために活用できる.
    \item 構成・再構成が容易であり, 構造を変化させられる.
  \end{itemize}
  これらの特徴を持った筋骨格ヒューマノイドプラットフォームを開発することで, 以下のような利点がある.
  \begin{itemize}
    \item 柔軟な身体と可変剛性制御により環境接触を含む学習制御等の模索を安全に行うことができる.
    \item 通常筋骨格ヒューマノイドは関節エンコーダを持たないが, 関節角度等を直接測定することで実験評価を容易にし, 開発を促進する.
    \item リンク長や身体自由度を自在に変えたり, 新しい粘弾性要素を加えたりすることで, 成長等の身体変化における自己身体像の変化について考察できる.
  \end{itemize}
}%

\switchlanguage%
{%
  To fulfill such requirements, we construct a new musculoskeletal humanoid Musashi using 3 concepts stated as below.
  \begin{itemize}
    \item A flexible body structure with nonlinear elastic elements having soft structures, and soft materials covering the folded back muscles.
    \item Joint modules which can directly measure joint angles and have compact sphere-shaped structures.
    \item A modularized body structure easily constructed using only joint modules, muscle modules, generic bone frames, muscle wire units, and a few attachments.
  \end{itemize}
  Musculoskeletal humanoids developed so far \cite{nakanishi2013design, wittmeier2013toward, jantsch2013anthrob, asano2016kengoro} have very complex bodies mimicking the detailed human musculoskeletal structure at the cost of design effort.
  By component modularization, this study aims to accelerate the investigation of learning control systems by increasing flexiblity, redundancy of internal sensors, and easiness of design, at the cost of human-like ball joints and human-like body proportion to some degree.
  % In the following sections, we will first state an overview of the proposed musculoskeletal structure, and detailed designs of the respective modules.
  % Next, we will state the design and system configuration of the simple musculoskeletal platform MusashiLarm and musculoskeletal humanoid platform Musashi, which extends MusashiLarm to the whole body.
  % Finally we will conduct several basic experiments and learning control experiments, and state the conclusion.
}%
{%
  これらの条件を満たすために, 以下の3つのコンセプトで筋骨格ヒューマノイドを構成する.
  \begin{itemize}
    \item それ自体が柔軟な非線形弾性ユニット, 筋の折り返しとソフトカバーによる柔軟身体構造
    \item 関節角度を直接測定可能かつコンパクトな球形状である関節モジュール
    \item 関節モジュール・筋モジュール・汎用骨格・筋経由点ユニット・少数のアタッチメントのみで簡易に構成可能なモジュラー身体構造
  \end{itemize}
  これまで開発されてきた筋骨格ヒューマノイド\cite{nakanishi2013design, wittmeier2013toward, jantsch2013anthrob, asano2016kengoro}は, 詳細な人体模倣のために設計コストを犠牲にし, 非常に複雑な身体を構成してきた.
  本研究では, 球関節や多少の身体プロポーションを犠牲にする代わりに全身をモジュール化し, 設計の容易さや柔軟性・センサの冗長性を増して, 筋骨格構造の学習型制御システム模索を促進することを至上命題とする.

  以降ではまず, 提案する筋骨格構造の概要を述べ, それを構成するそれぞれのモジュール設計について詳細に述べる.
  次に, 最もシンプルな構成である筋骨格プラットフォームMusashiLarm, それを全身まで拡張させた筋骨格ヒューマノイドプラットフォームMusashiの詳細な設計・システム構成について述べる.
  最後に, MusashiLarm・Musashiを用いたいくつかの予備実験と学習制御実験を行い, 結論を述べる.
}%

\begin{figure}[htb]
  \centering
  \includegraphics[width=0.9\columnwidth]{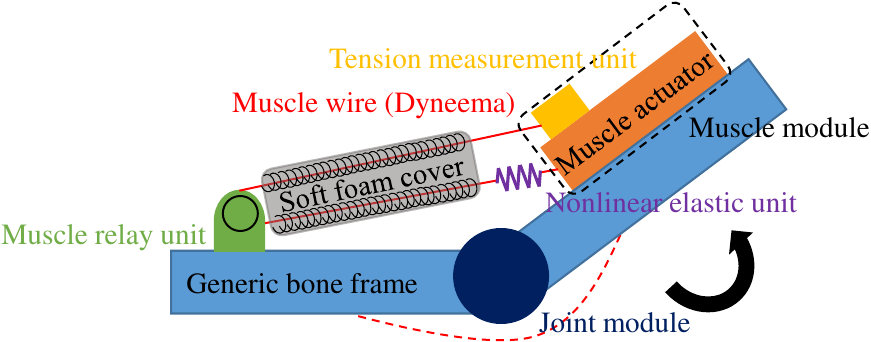}
  \caption{Overview of the proposed musculoskeletal structure.}
  \label{figure:musculoskeletal-overview}
  \vspace{-1.0ex}
\end{figure}

%%%%%%%%%%%%%%%%%%%%%%%%%%%%%%%%%%%%%%%%%%%%%%%%%%%%%%%%%%%%%%%%%%%%%%%%%%%%%%%%
\section{Development of Respective Modules} \label{sec:2}
\switchlanguage%
{%
  In this study, we propose the modularized musculoskeletal structure as shown in \figref{figure:musculoskeletal-overview}.
  The joint module, which can directly measure joint angles, connects generic bone frames together, and the muscle module is attached to the bone frame.
  The muscle wire goes out from the tension measurement unit attached to the muscle actuator, is folded back through the muscle relay unit, and is attached to the nonlinear elastic unit.
  Also, the folded back wire in each muscle is covered by a soft foam cover.
  There is a spring between them in order to inhibit the friction between the wire and soft foam cover.
  This muscle configuration inhibits the decrease of transmission efficiency due to the use of a motor with a high gear ratio, and we expect it to have the same effect as planar muscles \cite{osada2011planar} by covering the entire joint with folded back muscle wires.
  Also, by adding the nonlinear elastic units and soft foam covers, we enable soft environmental contact and variable stiffness control.

  % As the muscle modules, we use our previously developed sensor-driver integrated muscle modules \cite{asano2015sensordriver} and miniature bone-muscle modules \cite{kawaharazuka2017forearm}.
  In this study, we call the muscle components muscle wire unit, except for the muscle module.
  In the following sections, we will explain the detailed design of the joint modules, muscle modules, and muscle wire units.
}%
{%
  本研究では, \figref{figure:musculoskeletal-overview}のようなモジュール化された筋骨格構成を提案する.
  関節角度を直接測定可能な関節モジュールが骨格同士を繋ぎ, 骨格に対して筋モジュールが装着されている.
  筋アクチュエータに接続した筋張力測定ユニットから筋ワイヤが伸び, 筋経由点ユニットで折り返した終端部に非線形弾性ユニットを取り付けている.
  また, ワイヤの周りを柔らかい発泡性のカバーで覆っている.
  発泡性カバーとワイヤの摩擦を低減するため, ワイヤと発泡性カバーの間にはバネが介在している.
  この構成は, ギアの段数をあげることによる伝達効率の低下を防ぎ, かつ, 1つの筋において二本のワイヤが通るため, より関節全体を筋で覆え, 面状筋のような効果\cite{osada2011planar}が得られることが期待される.
  また, 折り返した終端部に非線形弾性要素を, 筋の周りに筋外装を取り付けることで, より柔らかい環境接触と可変剛性制御を可能とする.

  筋モジュールは, 先行研究で筆者らが開発したセンサドライバ統合型筋モジュール\cite{asano2015sensordriver}・骨構造一体小型筋モジュール\cite{kawaharazuka2017forearm}を用いる.
  本研究では, 全体の筋構成における筋モジュール以外の要素を総称して筋ワイヤユニットと呼ぶ.
  以降で, 関節モジュール・筋モジュール・筋ワイヤユニットの詳細な設計について述べる.
}%

\begin{figure}[htb]
  \centering
  \includegraphics[width=1.0\columnwidth]{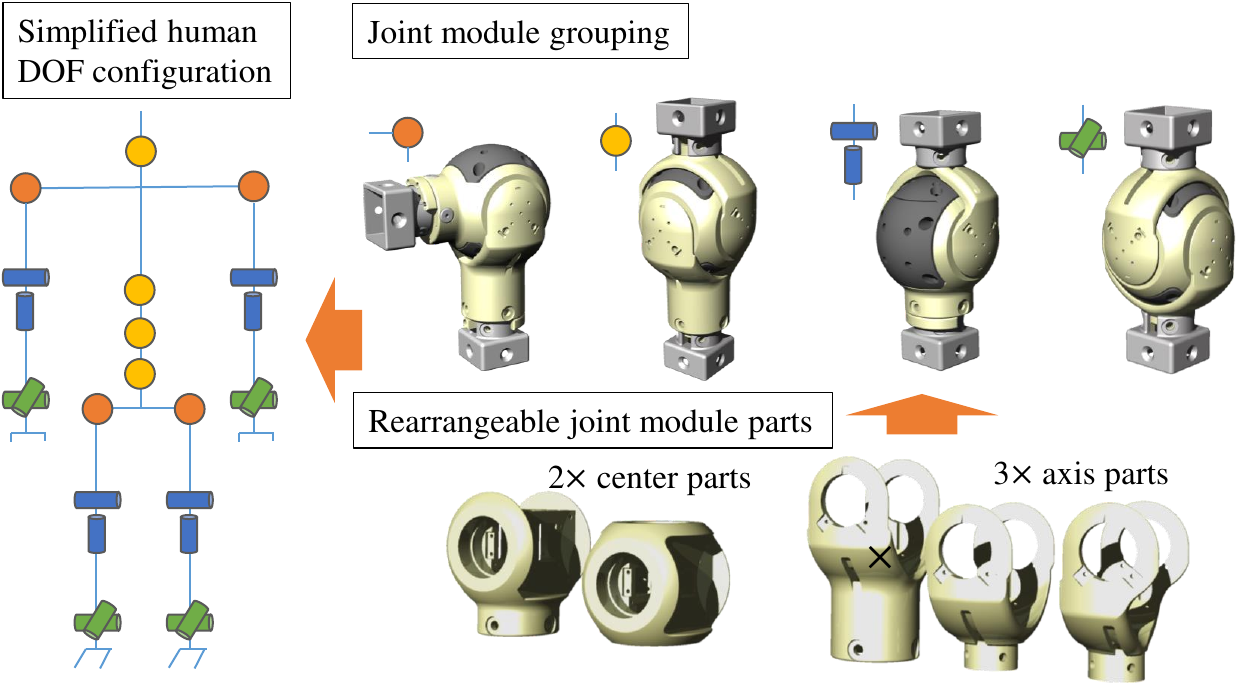}
  \vspace{-3.0ex}
  \caption{Grouping of simplified human joints, and application of joint modules.}
  \label{figure:jointmodule-grouping}
  \vspace{-1.0ex}
\end{figure}

\subsection{Joint Module}
\switchlanguage%
{%
  We set the requirements of the joint module, as below.
  \begin{enumerate}
    \item It has a spherical shape for the muscles to be able to cover the joint like in human beings.
    \item It can measure joint angles in some way.
    \item Its circuits and cables are packaged compactly, and it can function as a stand-alone unit.
    \item It has a general versatility and can be used as any joint of the musculoskeletal humanoid.
  \end{enumerate}

  Regarding (2), there are some methods to measure joint angles.
  In the ordinary axis-driven humanoid \cite{kaneko2002hrp}, each joint is independent, and each joint typically has an encoder or potentiometer.
  Also, there is a method to use an inertial measurement unit (IMU) to obtain joint angles from the acceleration and terrestrial magnetism.
  However, in the measurement of joint angles by IMU, we must use the terrestrial magnetism to inhibit the drift of yaw rotation, and so, it is difficult to measure joint angles correctly due to the noise of the many muscle motors included in the body of the musculoskeletal humanoid.
  Urata, et al. have developed a system to measure joint angles using a small camera \cite{urata2006sensor}, but we do not adopt it due to the system complexity.
  In this study, we design a sphere-shaped joint module which can directly measure joint angles using small potentiometers, like in the ordinary axis-driven humanoid.
  We compromise only on the point that it has a singularity like in the ordinary axis-driven humanoid.

  Regarding (3), we package the circuits and cables that read potentiometers into this joint module compactly, and make it possible to function as a stand-alone, only by connecting a USB cable.

  Regarding (4), the simplified human DoF configuration can be expressed as the left figure of \figref{figure:jointmodule-grouping}, if we ignore the complex scapula joint.
  We can group the joints into just 4 kinds of joint modules as shown in the upper right figure of \figref{figure:jointmodule-grouping}.
  We construct these 4 kinds of joint modules by combination of the 2 kinds of center parts and 3 kinds of axis parts, as shown in the lower right figure of \figref{figure:jointmodule-grouping}.
  By combining these 5 rearrangeable parts, we can construct various joint modules other than the 4 kinds of joint modules by adding or removing DoFs.
}%
{%
  筋骨格ヒューマノイドの関節として適用でき, 関節角度を測定することのできる関節モジュールの要件を本研究では以下のように設定した.
  \begin{enumerate}
    \item 人間のように筋肉が関節を覆うことができるよう, 関節は球形である.
    \item 各関節角度を直接測定することができる.
    \item コンパクトに回路等を含んでパッケージ化されており, 単体として簡易に動作する.
    \item 汎用性があり, 筋骨格ヒューマノイドの全関節をこの関節モジュールで担うことができる.
  \end{enumerate}
  (2)に関してはいくつかの方法が考えられる.
  通常のヒューマノイドであればそれぞれの関節は独立しており, それぞれエンコーダやポテンショメータがついた構成が一般的である\cite{kaneko2002hrp}.
  また, IMUを用いて加速度と地磁気から関節角度を測る方法も存在するが, yaw軸回転のドリフトを抑えるため地磁気を用いており, 冗長な多数の筋のモータが体中に存在する筋骨格ヒューマノイドに適用するのは難しい.
  また, Urataらは球関節の角度を小さなカメラを用いて測定する優れたシステムを開発しているが\cite{urata2006sensor}, システムの複雑性等から本研究では採用しない.
  本研究では, 球関節という形を取りながらも, その中に収納できるよう小さなポテンショメータを用いて通常の軸駆動型ヒューマノイドと同様の構造で関節角度を直接測定可能な擬似球関節モジュールの設計を行う.
  しかし, 通常の軸駆動ヒューマノイドの関節と同様, 特異点が生まれてしまうという問題のみ, 本研究では妥協する.
  (3)については, ポテンショメータを読む回路を擬似球関節モジュールの中に内包しパッケージ化することで, 単体としてUSBを繋ぐだけで動作するコンパクトなモジュール化を行う.
  (4)について, 人間の関節は, 複雑な肩甲骨関節等を除けば4種類の関節のみによって, シンプルに\figref{figure:jointmodule-grouping}の左図のように表せる.
  これらの主要な関節群を少数の部品の組み換えによって全て表現可能な関節モジュールを作ることを考える.
  そこで, \figref{figure:jointmodule-grouping}の右図のように, これら4種類の関節モジュールを, 関節モジュールの核となる2種類の中心パーツと, 3種類の回転パーツを複数組み合わせることで構築することができると考えた.
  また, この5種類のパーツを組み替えることで, ある自由度を消したり, 増やしたり等, この4種類以外にも様々な構成の関節モジュールを構成することが可能である.
}%

\begin{figure}[htb]
  \centering
  \includegraphics[width=0.9\columnwidth]{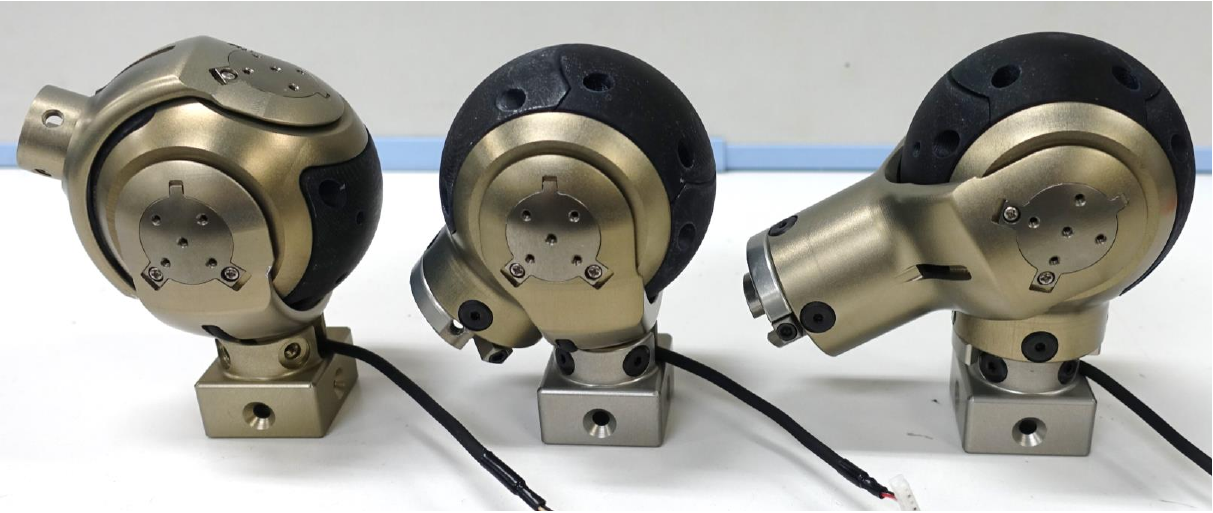}
  \caption{Overview of the newly developed joint modules.}
  \label{figure:jointmodule-overview}
  \vspace{-1.0ex}
\end{figure}

\begin{figure*}[htb]
  \centering
  \includegraphics[width=1.7\columnwidth]{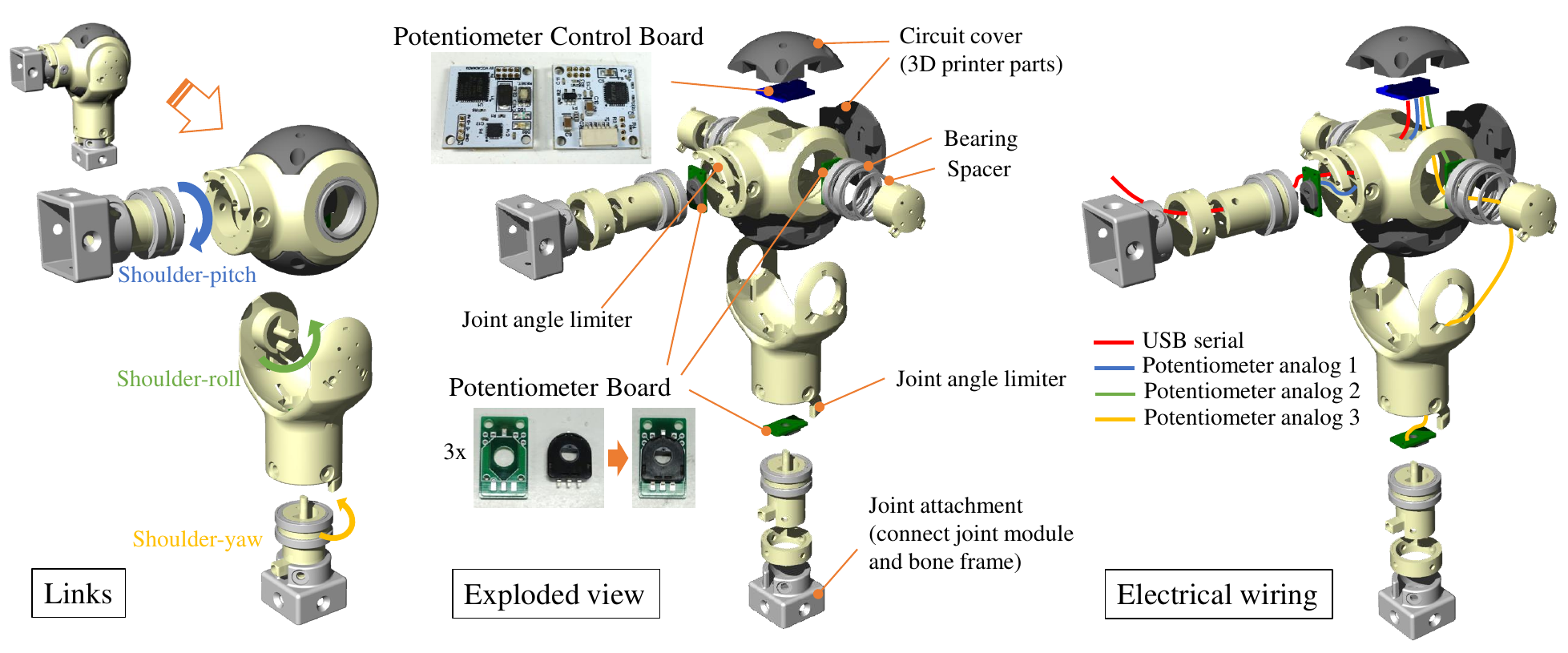}
  \caption{Detailed design of the joint module, for the shoulder joint.}
  \label{figure:jointmodule-detail}
  \vspace{-3.0ex}
\end{figure*}

\switchlanguage%
{%
  We show the developed joint modules in \figref{figure:jointmodule-overview}.
  In order from the left, the joint modules are for the wrist, elbow, and shoulder joints.

  The detailed structure of this joint module is shown in \figref{figure:jointmodule-detail}.
  This is the structure of the shoulder joint module, and it is divided into 4 links, and 3 rotation axes exist among these links.
  Its structure is made by cutting A7075, and other parts are bearings, spacers, and circuit covers.
  A Potentiometer Board which reads the output of a potentiometer is arranged at each joint axis, and a Potentiometer Control Board which integrates them is arranged in the joint module.
  Also, its electrical wiring is shown in the right figure of \figref{figure:jointmodule-detail}.
  All of the electrical wiring that connects the Potentiometer Control Board and Potentiometer Boards passes through the joint module, except for one USB cable, which connects the joint module to the outside PC.
  The Potentiometer Control Board is equipped with IMU, which can be used for a more correct measurement of postures.
}%
{%
  実際に開発された関節モジュールを\figref{figure:jointmodule-overview}に示す.
  \figref{figure:jointmodule-overview}にあるのは, 左から, 手首, 肘, 肩に用いる関節モジュールである.

  関節モジュールの構造の詳細は\figref{figure:jointmodule-detail}となっている.
  これは肩の関節モジュールの構造であるが, 基本的には4つのリンクに分かれており, その間に3つの回転軸が存在する.
  構造材はA7075を使用して切削されており, 他の素材の部品としてはベアリング・スペーサ・3Dプリンタで作成された回路保護カバーが存在する.
  ポテンショメータを読むPotentiometer Boardはそれぞれの関節軸ごとに配置されており, それらを統合するPotentiometer Control Boardが1つの関節モジュールに必ず1つ存在し, モジュール内部に収納されている.
  また, ケーブルの配線は\figref{figure:jointmodule-detail}の右図にあるような配線となっている.
  Potentiometer Control BoardとPotentiometer Boardを繋ぐ配線は全て関節モジュールの中を通っており, 外部とUSB通信を行うためのケーブルが1本だけ関節モジュールから外に出ている.
  Potentiometer Control BoardにはIMUが搭載されており, 並進加速度・回転角速度・地磁気の9軸の値を得ることができる.
}%

\subsection{Muscle Module}
\begin{figure}[htb]
  \centering
  \includegraphics[width=1.0\columnwidth]{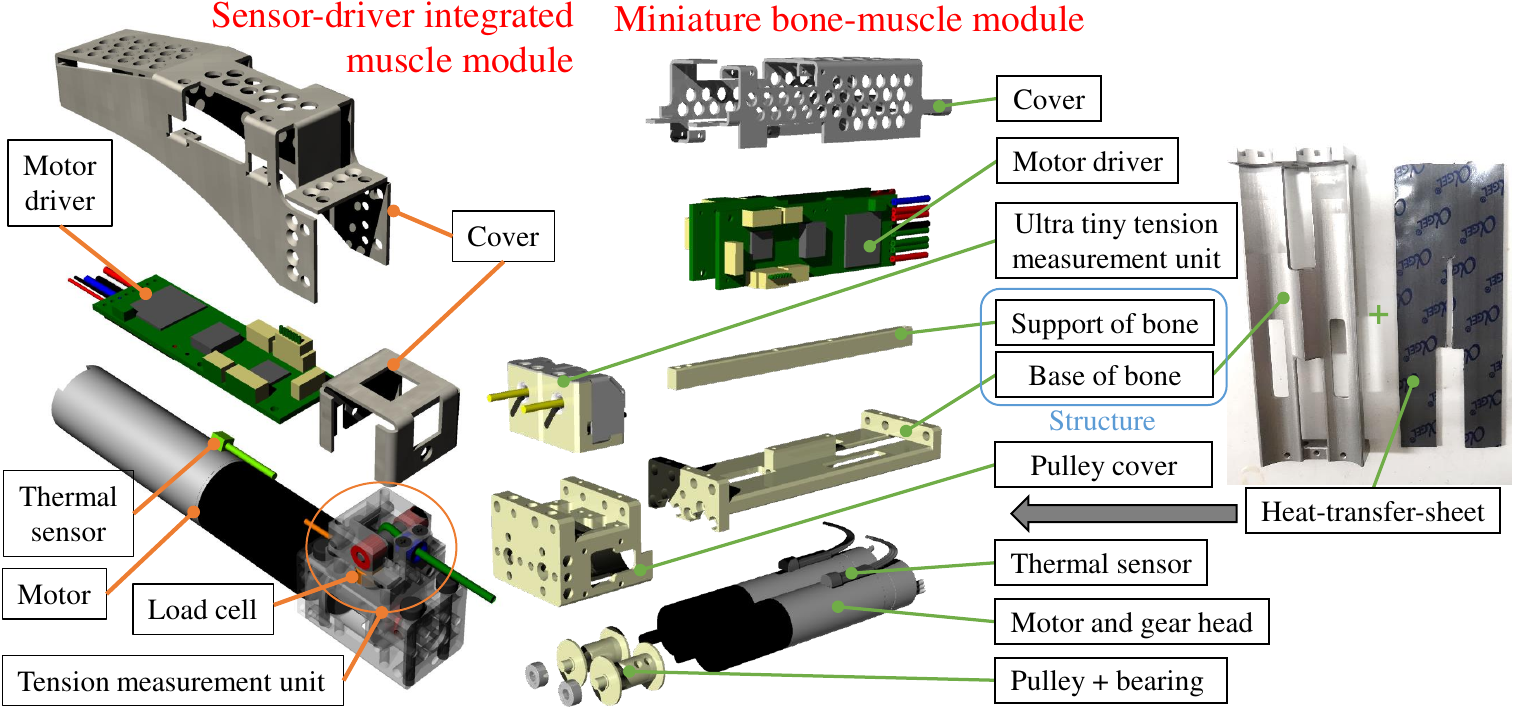}
  \vspace{-3.0ex}
  \caption{Detailed design of the sensor-driver integrated muscle module \cite{asano2015sensordriver} and miniature bone-muscle module \cite{kawaharazuka2017forearm}.}
  \label{figure:muscle-module}
  \vspace{-1.0ex}
\end{figure}

\begin{figure}[htb]
  \centering
  \includegraphics[width=1.0\columnwidth]{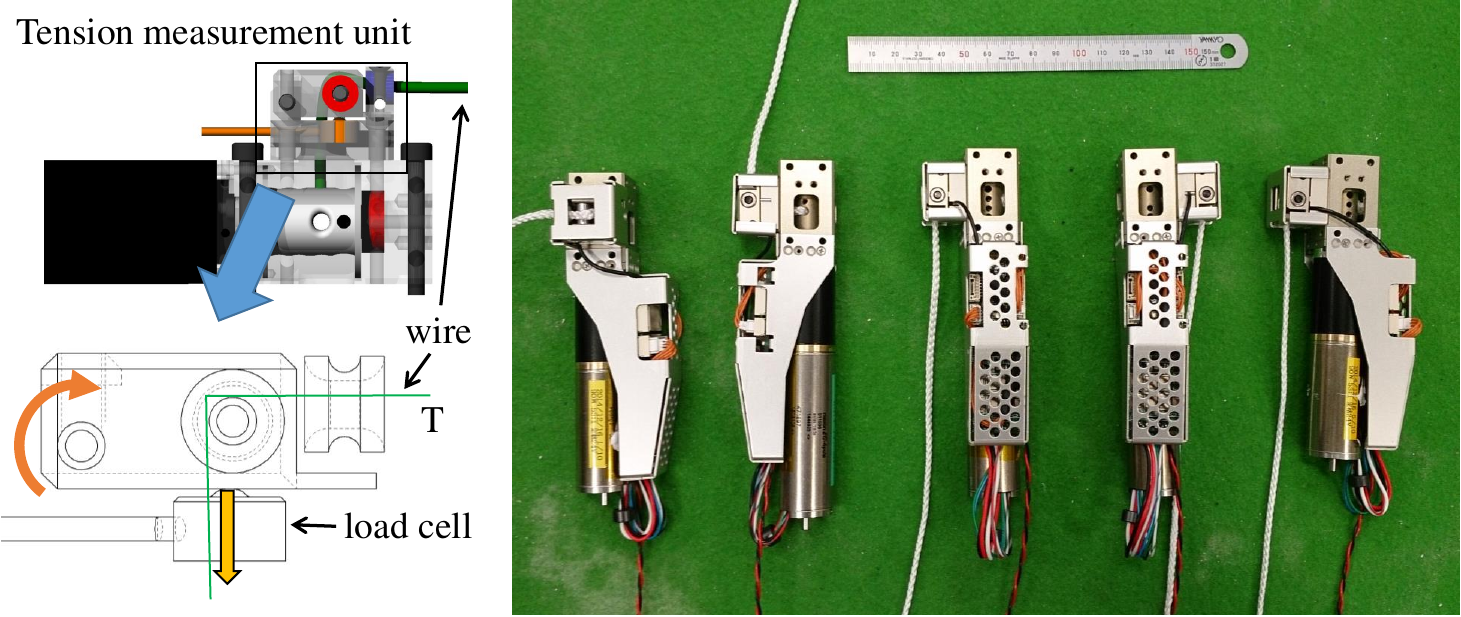}
  \vspace{-3.0ex}
  \caption{The structure of the tension measurement unit, and the versatility of its arrangements to the muscle actuator.}
  \label{figure:tension-unit}
  \vspace{-1.0ex}
\end{figure}

\switchlanguage%
{%
  First, we show the details of the sensor-driver integrated muscle module \cite{asano2015sensordriver} in the left of \figref{figure:muscle-module}.
  This is the muscle module which dramatically increases the modularity and reliability by including $\phi22$ brushless DC motor, a pulley winding the muscle wire, motor driver, thermal sensor, tension measurement unit, circuit cover, etc. into one package.
  We can freely change the gear ratio of the actuator, but we used 29:1 or 53:1 in this study.
  The motor driver can conduct a current control and muscle tension control feedbacking muscle tension.
  We use chemical fiber Dyneema with high abrasion resistance as the muscle wires.
  The structure of the tension measurement unit is shown in the left of \figref{figure:tension-unit}, and this unit can measure muscle tension up to 500 N by converting it to the pressure to the loadcell using the angular moment around the axis.
  Also, as shown in the right of \figref{figure:tension-unit}, the tension measurement unit can be attached to the muscle actuator from various directions and realize various muscle routes.

  Next, we show the details of the miniature bone-muscle module \cite{kawaharazuka2017forearm} in the right of \figref{figure:muscle-module}.
  A major difference of this module compared to the previous module is that this module can be used as not only the muscle actuator but also as the bone structure.
  This module includes 2 small $\phi16$ brushless DC motors.
  By including multiple actuators in one module, we can use space efficiently.
  ``Base of bone'' and ``Support of bone'' in \figref{figure:muscle-module} become the body structure, and we can compactly construct the body structure with only muscle modules, by connecting these modules lengthwise and crosswise.
  Also, this module can compensate for the drawbacks adopting these small actuators, by dissipating the heat of the muscle actuator to ``Base of bone'' through ``Heat-transfer-sheet''.

  These muscle modules are excellent in terms of the modularity, reliability, and versatility, so they can be the foundation of the musculoskeletal platform.
}%
{%
  まず, センサドライバ統合型筋モジュールの詳細を\figref{figure:muscle-module}の左図に示す.
  これは, アクチュエータとなる$\phi22$のBLDC Motor・筋を巻き取るプーリ・モータドライバ・温度センサ・筋張力測定ユニット・回路カバー等を一つのパッケージの中に収めることで, モジュール性・信頼性を格段に上げた筋モジュールである.
  アクチュエータはギア比を自由に変更することができ, 本研究では基本的に29:1か53:1を用いる.
  モータドライバは電流制御が可能であり, 筋張力をフィードバックループに組み込んだ筋張力制御を行うことができる.
  筋には摩擦に強い化学繊維であるDyneemaを使用している.
  筋張力測定ユニットの構造は\figref{figure:tension-unit}の左図のようになっており, 軸中心の回転モーメントを使い, 筋の張力をロードセルの圧力に変換することで500 Nまでの筋張力を測定できる.
  また\figref{figure:tension-unit}の右図のように, 筋アクチュエータに対して様々な方向から筋張力測定ユニットを取り付けることができ, 様々な筋経路を実現できるという汎用性を持つ.

  次に, 骨構造一体小型筋モジュールの詳細を\figref{figure:muscle-module}の右図に示す.
  この筋モジュールが先の筋モジュールと最も異なる点は, この筋モジュール自体を骨格として用いることができる点である.
  アクチュエータはセンサドライバ統合型筋モジュールのアクチュエータより一回り小さな$\phi16$のBLDC motor 2本を使用している.
  複数のアクチュエータを一つのモジュール内に収納することで, 効率的な空間配置を取ることができる.
  ``Base of bone''と``Support of bone''が筋モジュールの骨格となり, この筋モジュールを縦や横に接続することで, 筋モジュールのみでコンパクトに骨格を構成できる点において優れている.
  また, 小さなアクチュエータを使うことによる筋張力のビハインドを, ``Heat-transfer-sheet''を介してアクチュエータの熱を``Base of bone''に逃がすことで補っている.

  これら筋モジュールは, モジュール化という観点で優れており, 信頼性・汎用性が高いことから, 筋骨格プラットフォームのアクチュエータ基盤となる.
}%

\subsection{Muscle Wire Unit}
\subsubsection{Muscle Relay Unit}
\begin{figure}[htb]
  \centering
  \includegraphics[width=0.8\columnwidth]{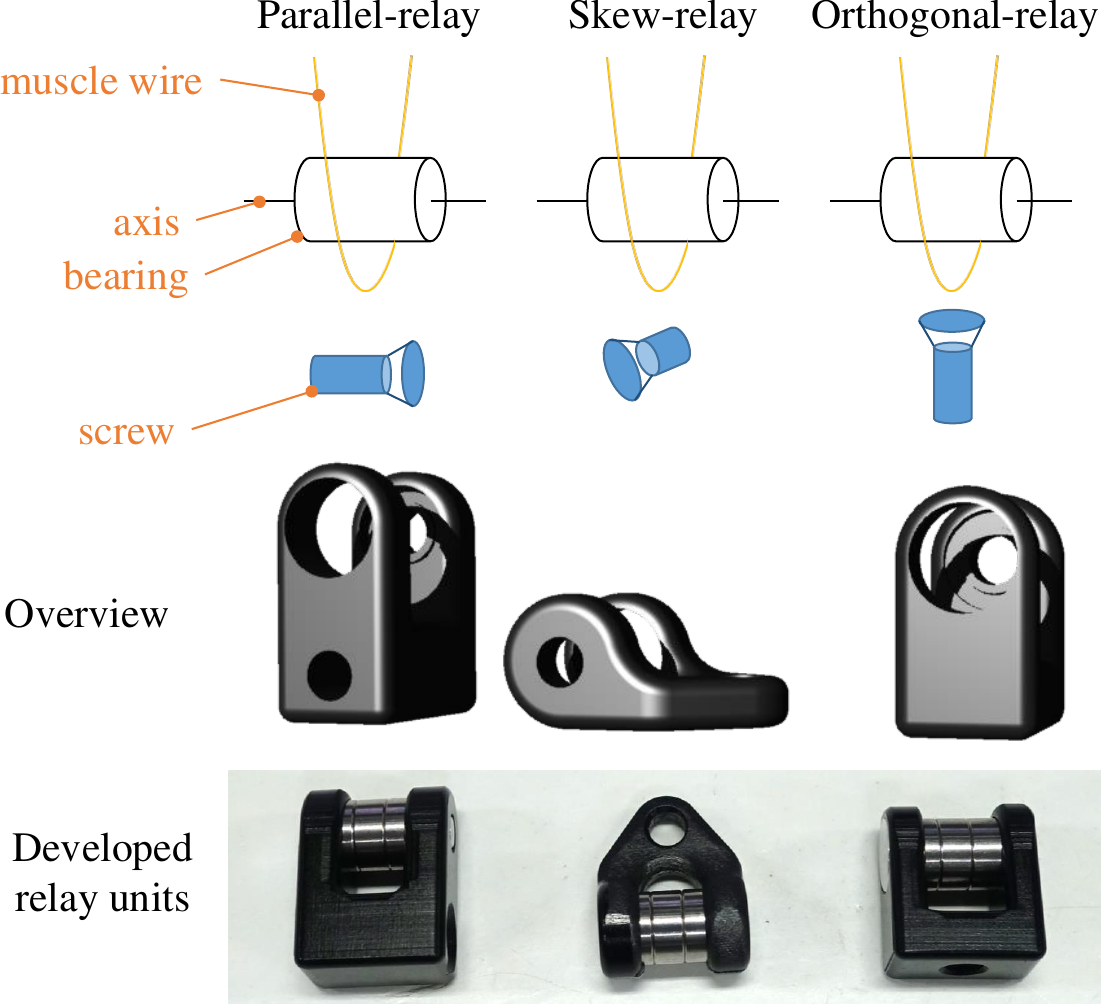}
  \caption{Standardized 3 muscle relay units: Parallel-relay, Skew-relay, and Orthogonal-relay.}
  \label{figure:muscle-relay}
  \vspace{-1.0ex}
\end{figure}

\switchlanguage%
{%
  We develop the muscle relay unit, which folds back a muscle wire, by arranging an axis with 3 bearings into a base structure.
  We can realize compact muscle relay units by directly using bearings to fold back muscles, without embedding the bearing into the structure.
  The structure of the muscle relay units can be standardized by the direction of attachment to the bone frame and the direction of the folded back muscle.
  We need to design only 3 kinds of muscle relay units shown in the upper figure of \figref{figure:muscle-relay}.
  We show the developed muscle relay units in the lower figure of \figref{figure:muscle-relay}.
  This is the result of considering the most compact shape when setting the direction of the axis and the attachment by a screw.
  By choosing these units from the direction of the tapped holes and the direction the muscle will be folded back, we can quickly realize the muscle relay mechanism.
}%
{%
  筋経由点ユニットは, ある一つのベースとなる構造に対して, 周りに3つの連続したベアリングを備える一本の軸を配置した構造として開発している.
  ベアリングをベースの構造に対して組み込まず, 直接筋を折り返す場所に使うことで, コンパクトな筋経由点ユニットを実現している.
  筋経由点ユニットの構造は, それを取り付けるネジの方向と, 筋の折り返しの方向の違いによって, \figref{figure:muscle-relay}の上図のような3種類のみ作成すれば事足りる.
  \figref{figure:muscle-relay}の下図に開発した筋経由点ユニットを示すが, これらは一本のネジの方向と, 連続したベアリングつきの軸の方向が定まった際に, その中で最もコンパクトな形状を考えた結果である.
  この3種類の筋経由点ユニットを, 骨格のネジ穴の方向と, 折り返したい筋の方向から選定し, それぞれ適用することで, 筋折り返し構造を実現可能である.
}%

\subsubsection{Nonlinear Elastic Unit}
\begin{figure}[htb]
  \centering
  \includegraphics[width=1.0\columnwidth]{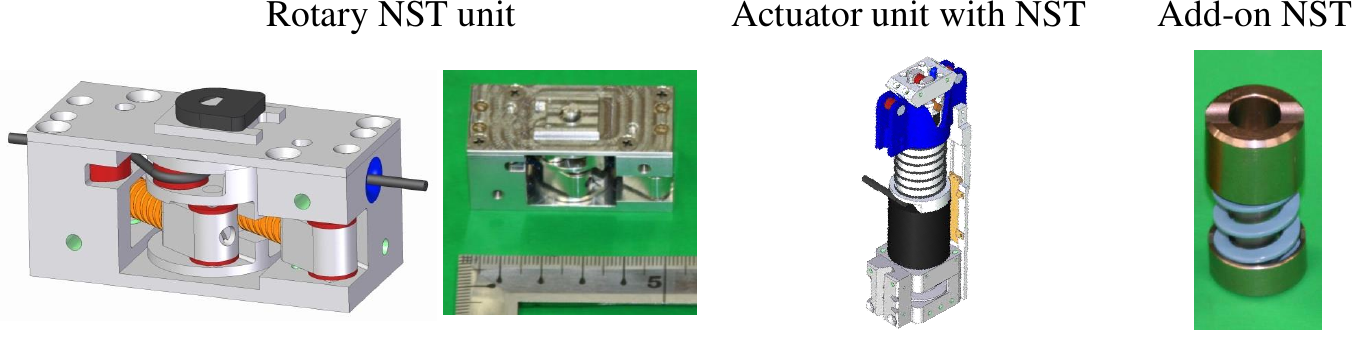}
  \vspace{-3.0ex}
  \caption{Previous nonlinear spring tensioner units \cite{nakanishi2011kenzoh, osada2010addon}.}
  \label{figure:previous-nonlinearunit}
  \vspace{-1.0ex}
\end{figure}

\begin{figure}[htb]
  \centering
  \includegraphics[width=0.95\columnwidth]{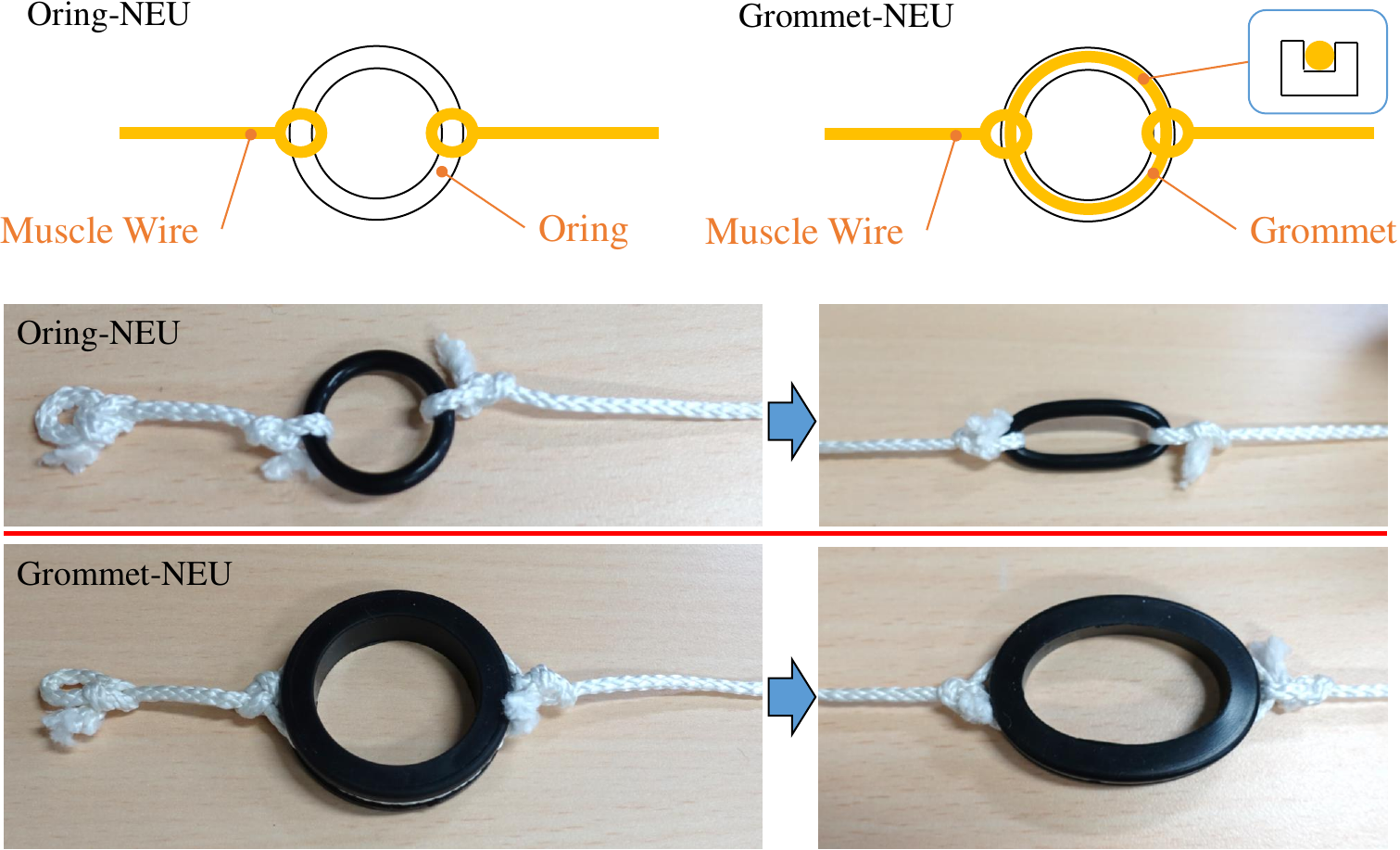}
  \caption{Developed nonlinear elastic units: Oring-NEU and Grommet-NEU.}
  \label{figure:current-nonlinearunit}
  \vspace{-1.0ex}
\end{figure}

\switchlanguage%
{%
  We show the previously developed nonlinear spring units in \figref{figure:previous-nonlinearunit}.
  The two units in the left of \figref{figure:previous-nonlinearunit} are the nonlinear spring tensioner (NST) units for Kenzoh \cite{nakanishi2011kenzoh}, and the right unit is the add-on nonlinear spring unit \cite{osada2010addon}.
  Although these units can geometrically realize nonlinear elasticity, they are too hard and large to use for the human-mimetic musculoskeletal structure because they are generally composed of metal and springs.

  In this study, we develop a compact and flexible nonlinear elastic unit (NEU).
  The closest way to realize it is the method using O-ring, included in Anthrob \cite{jantsch2013anthrob}.
  However, the elastic unit in \cite{jantsch2013anthrob} could not express nonlinear elasticity.

  We show the developed nonlinear elastic unit in \figref{figure:current-nonlinearunit}.
  The first unit is Oring-NEU.
  Although this is basically the same as the unit in Anthrob \cite{jantsch2013anthrob}, only nitrile rubber could express nonlinear elasticity, when we tried some materials of O-ring (urethane, silicon, ethylene propylene, fluorine, and nitrile).
  However, nitrile rubber deteriorates by ozone in the air, and is easily ruptured from cracks.
  Also, because nonlinear elasticity can be obtained by pulling on the O-ring, there is a limit of muscle tension due to the strength of the rubber.

  Then, we developed Grommet-NEU.
  The nonlinear elasticity can be expressed by compressing the grommet structure using the rolled muscle wire around it.
  In this structure, we can geometrically realize nonlinear elasticity, and so can freely choose its material.
  Also, the strength increases dramatically compared to Oring-NEU, because only the compression of the rubber is used.
  In this study, we use weather and water resistant ethylene propylene rubber as the grommet.
}%
{%
  これまで開発されたきた非線形弾性ユニットを\figref{figure:previous-nonlinearunit}に示す.
  \figref{figure:previous-nonlinearunit}の左２つは\cite{nakanishi2011kenzoh}に用いられたNSTであり, 一番右は長田らによって開発された, 腱に対して後から装着することができるアドオンユニットである\cite{osada2010addon}.
  これらは幾何的に非線形弾性要素を実現できる一方, 基本的に金属とばねで構成されており, 大きく, 硬い構造をしているため, 人体模倣型筋骨格ヒューマノイドの身体への適用は難しい.

  本研究では, コンパクトで柔軟な非線形弾性要素を実現するための, 非線形弾性ユニットの構造について考える.
  最もそれに近い構造は\cite{jantsch2013anthrob}でも用いられている, O-ringを使った方法であろう.
  しかし, \cite{jantsch2013anthrob}のユニットでは非線形弾性を表現しきれていない.

  \figref{figure:current-nonlinearunit}に本研究で開発した非線形弾性ユニットの詳細を示す.
  一つ目はOring-NEUであり, これは基本的に\cite{jantsch2013anthrob}で用いられていたものと同じであるが, いくつかの素材を試し, ニトリルゴムのみが正しく非線形弾性を表現することができた.
  しかし, ニトリルゴムは大気中に含まれるオゾンや湿気, 及び温度によって劣化を引き起こし, クラックが入ることでゴム自体が切れやすくなってしまうという問題があった.
  また, この構造はゴムを引き伸ばすことによって非線形弾性を得るため, ゴムの強度によって筋張力に限界が生じてしまう.

  そこで, ２つ目のGrommet-NEUを開発した.
  これは, 筋のワイヤによってグロメット構造を圧縮することで非線形弾性を表現する.
  この構造においては非線形弾性を幾何的に実現できるため, 素材を自由に選ぶことができる.
  また, Oring-NEUと違い, ゴムの圧縮のみを用いるため強度が格段に向上している.
  本研究では耐候性や耐水性のあるエチレンプロピレンゴムを用い, 非線形弾性要素を実現している.
}%

\begin{figure}[t!]
  \centering
  \includegraphics[width=1.0\columnwidth]{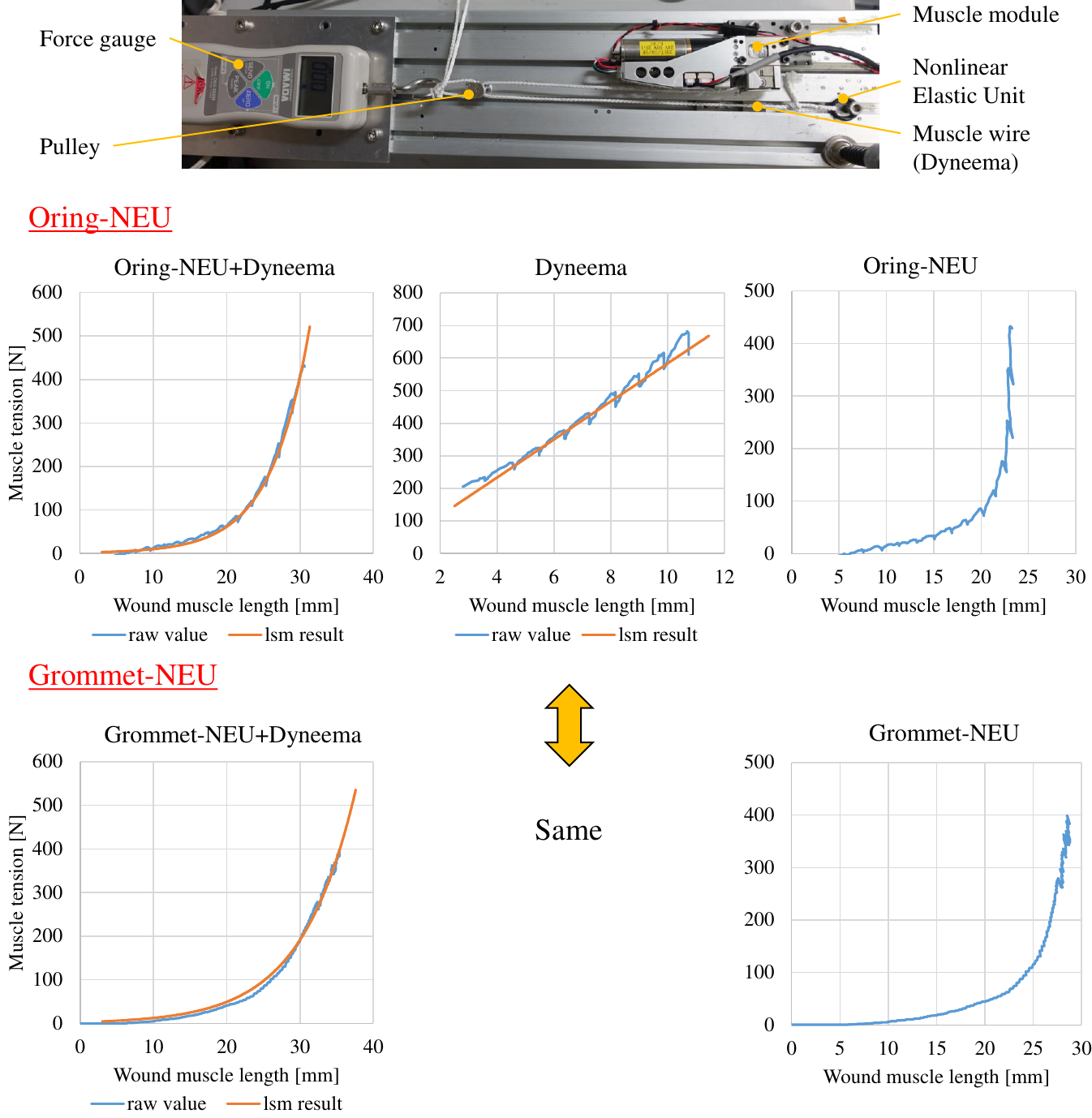}
  \vspace{-3.0ex}
  \caption{Identification of nonlinear elastic parameters.}
  \label{figure:nonlinear-measurements}
  \vspace{-1.0ex}
\end{figure}

\switchlanguage%
{%
  Finally, we conducted a parameter identification of this muscle configuration to verify the nonlinear elastic feature.
  The result is shown in \figref{figure:nonlinear-measurements}.
  As shown in the upper figure of \figref{figure:nonlinear-measurements}, we fix the absolute muscle length, wind the muscle wire, and identify the parameter by examining the amount of the wound muscle and the value of a force gauge.
  We assume that the relative change in muscle length is sufficiently small compared to the absolute muscle length.
  We show the result of the changes in muscle length and muscle tension in this condition, with the nonlinear elastic unit in the left figure, without the nonlinear elastic unit in the center figure, and with only the nonlinear elastic unit in the right figure of \figref{figure:nonlinear-measurements}.
  In this experiment, we fixed the absolute muscle length to 480 mm, and arranged the origin of the graphs by using the assumption that the elongation of the muscle is 0 when the muscle tension is 0.
  We can estimate the feature of the Dyneema as shown below, by using the assumption that its elongation is linear to the muscle tension and its spring constant is inversely proportional to the absolute muscle length, and executing least-square method (LSM):
  \begin{align}
    T=a_{d}\Delta{l}_{d}/l_{abs} \label{eq:dyneema}
  \end{align}
  where $T$ [N] is the muscle tension obtained from the force gauge, $a_{d}$ is the spring constant of Dyneema per unit length (in this study, $a_{d}$ is estimated to be $2.8E+4$ from the lower center figure of \figref{figure:nonlinear-measurements}), $\Delta{l}_{d}$ [mm] is the muscle elongation, and $l_{abs}$ [mm] is the absolute muscle length.
  Also, we can estimate the parameters of this muscle configuration by LSM using an exponential function as shown below:
  \begin{align}
    T=a_{m}\textrm{exp}(b_{m}\Delta{l}_{m}) \label{eq:sum}
  \end{align}
  where $a_{m}, b_{m}$ are constants which express the nonlinear elastic feature (in this study, we estimated that $a_{m}=1.34, b_{m}=0.19$ regarding Oring-NEU, and $a_{m}=3.32, b_{m}=0.14$ regarding Grommet-NEU, from the left figure of \figref{figure:nonlinear-measurements}).
  By using \equref{eq:dyneema} and \equref{eq:sum}, the feature of the nonlinear elastic unit is expressed as shown below:
  \begin{align}
    \Delta{l}_{o}=\Delta{l}_{m}-\Delta{l}_{d}=\frac{1}{b_{m}}\textrm{log}(\frac{T}{a_{m}}+1)-\frac{Tl_{base}}{a_{d}} \label{eq:oring}
  \end{align}
  where $l_{base}$ is the absolute muscle length of 480 mm that is fixed in this study.
  So, by summing up the effects of \equref{eq:dyneema} and \equref{eq:oring}, we can estimate the entire feature of this muscle configuration.
  $T$ used in these equations is the output of the force gauge, and in actual use, we can measure half of $T$ from the tension measurement unit, because each muscle is folded back through a pulley.
  Also, the muscle features of the actual robot may be different from the estimated features due to the effects of the soft foam cover, friction, etc., so the actual elastic feature needs to be learned by using the sensor information of the actual robot \cite{kawaharazuka2018bodyimage}.
}%
{%
  最後に, これらの非線形弾性要素を用いた筋構成の非線形弾性パラメータの同定を行った結果を\figref{figure:nonlinear-measurement}に示す.
  \figref{figure:nonlinear-measurement}の上図のように, 絶対筋長を固定し, 筋を巻き取り, 筋が巻きとられた量・フォースゲージの力の値を調べることで, パラメータを同定する.
  筋の絶対長さに対して筋長の相対的な変化は十分小さいものと仮定する.
  \figref{figure:nonlinear-measurement}の左図にはこの筋構成全体の筋長変化・筋張力変化のグラフ, 中図には非線形弾性要素を外した際の結果を示している(なお, 絶対筋長は480 mmに固定しており, 筋張力0のときに筋の伸びは0という仮定から原点を揃えている).
  このとき, Dyneemaの伸びは線形, Dyneemaのばね定数は絶対筋長に反比例するという仮定を用いて最小二乗法を行うと, Dyneemaの特性は以下のように推定できる.
  \begin{align}
    T=a_{d}\Delta{l}_{d}/l_{abs} \label{eq:dyneema}
  \end{align}
  ここで, $T$ [N]はフォースゲージから得た筋張力, $a_{d}$は単位長さ辺りのDyneemaのばね定数, $\Delta{l}_{d}$ [mm]は筋の伸び, $l_{abs}$ [mm]は筋全体の絶対長さであり, 本推定では$a_{d}=2.8E+4$と推定された.
  また, この筋構成全体の特性は指数関数による最小二乗法によって以下のように求まる.
  \begin{align}
    T=a_{m}\exp(b_{m}\Delta{l}_{m}) \label{eq:sum}
  \end{align}
  ここで, $a_{m}, b_{m}$は非線形特性を表す定数であり, 本推定ではOring-NEUにおいて$a_{m}=1.34, b_{m}=0.19$, Grommet-NEUにおいて$a_{m}=3.32, b_{m}=0.14$と推定された.
  \equref{eq:dyneema}と\equref{eq:sum}を用いることで, 非線形弾性要素の特性は以下のようになる.
  \begin{align}
    \Delta{l}=\Delta{l}_{m}-\Delta{l}_{d}=\frac{1}{b_{m}}\log(\frac{T}{a_{m}}+1)-\frac{Tl_{base}}{a_{d}} \label{eq:oring}
  \end{align}
  ここで, $l_{base}$は今回の推定で固定した筋全体の絶対筋長の480 mmである.
  よって, \equref{eq:oring}と\equref{eq:dyneema}の効果を足し合わせることで, 本研究で用いる筋構成の特性がわかる.
  本推定で用いている$T$はフォースゲージで測れる張力であり, プーリを介して折り返しているため, 実際には張力センサユニットからはこの$T$の半分の力が計測される.
  また, 実際には, 発泡性カバーや摩擦等の影響により, 実機とは異なることがあり, 実機による非線形パラメータの学習をする必要がある\cite{kawaharazuka2018bodyimage}.
}%

%%%%%%%%%%%%%%%%%%%%%%%%%%%%%%%%%%%%%%%%%%%%%%%%%%%%%%%%%%%%%%%%%%%%%%%%%%%%%%%%
\section{Design of Modularized Musculoskeletal Humanoids} \label{sec:3}

\subsection{MusashiLarm: Modularized Musculoskeletal Upper Limb}
\subsubsection{Body Structure}
\begin{figure}[htb]
  \centering
  \includegraphics[width=1.0\columnwidth]{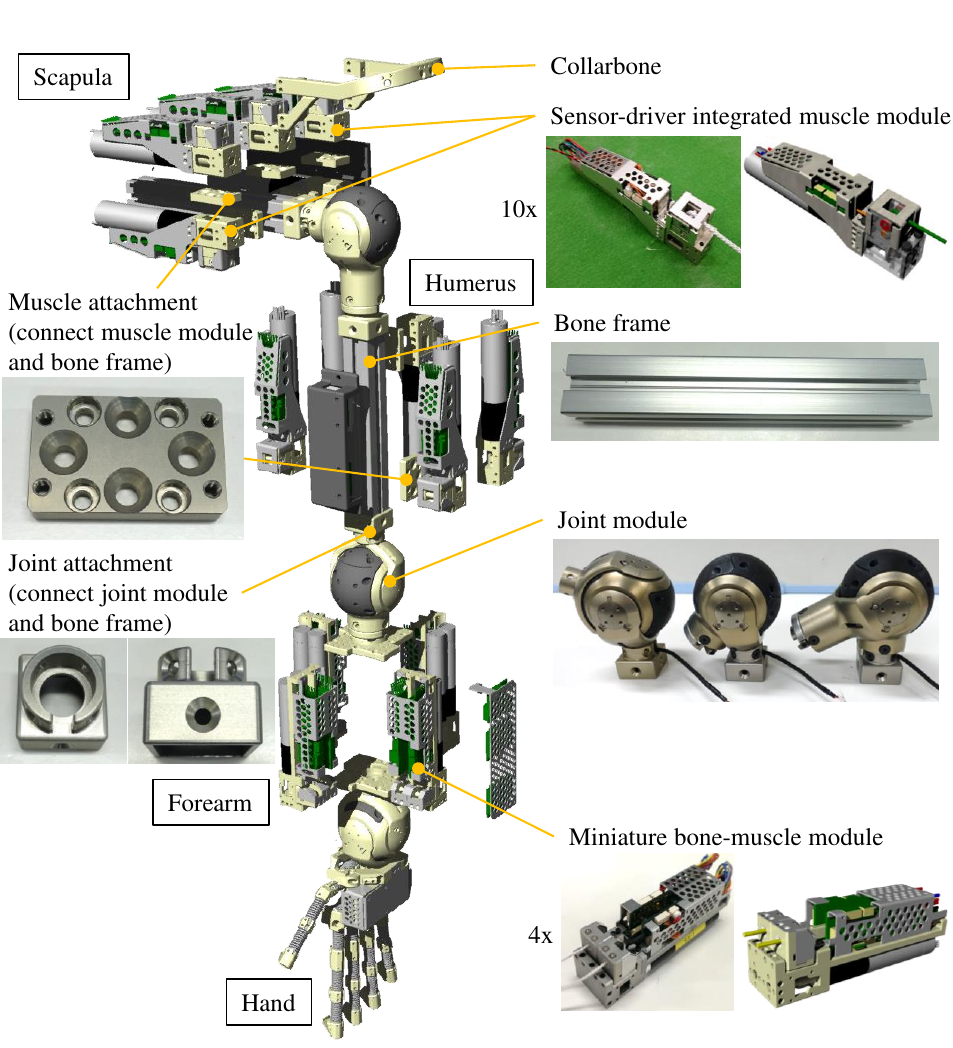}
  \vspace{-3.0ex}
  \caption{Detailed modular design of MusashiLarm.}
  \label{figure:musashilarm-detail}
  \vspace{-1.0ex}
\end{figure}

\switchlanguage%
{%
  The detailed design of MusashiLarm is shown in \figref{figure:musashilarm-detail}.
  MusashiLarm is composed of 4 links and 3 joints.
  We call these links the scapula, humerus, forearm and hand, respectively.
  The links of the scapula and humerus are constructed by aluminum generic bone frames, and in each link, 5 sensor-driver integrated muscle modules \cite{asano2015sensordriver} are each connected to the generic bone frame through a muscle attachment.
  We can connect the muscle module to the generic bone frame, and the muscle modules to each other, through the muscle attachment.
  Also, the joint module and generic bone frame are connected through a joint attachment, and if we change the joint attachment, we can change the configuration of the robot easily.
  The link of the forearm is composed of 4 miniature bone-muscle modules \cite{kawaharazuka2017forearm} which can connect lengthwise and crosswise as a bone structure.
  By using the benefit that the module can not only act as muscles but also as a bone structure, we do not need to use the bone frame, and the forearm link can easily be constructed using only the muscle modules.
  Also, the hand has flexible finger joints using machined springs \cite{makino2018hand}.

  In this way, we can realize easily constructable and reconstructable musculoskeletal body structures using joint modules, muscle modules, muscle wire units, generic bone frames, and a few attachments (muscle and joint attachments).
}%
{%
  開発したMusashiLarmの設計詳細を\figref{figure:musashilarm-detail}に示す.
  本研究で開発した筋骨格上肢は基本的に4つのリンク・3つの関節によって構成されている.
  それぞれのリンク名をScapula, Humerus, Forearm, Handと呼ぶこととする.
  Scapula・Humerusリンクの構造体はアルミの汎用骨格(Bone frame)によって構成されており, その周りに各リンクごとに5つのセンサドライバ統合型筋モジュール(Sensor-driver integrated muscle module)が筋アタッチメント(Muscle attachment)を介して装着されている.
  Muscle attachmentを介して, Bone frameとMuscle module, また, Muscle moduleとMuscle module同士を合体させることが可能である.
  また, 関節モジュール(Joint module)とBone frameはJoint attachmentを介して接続されており, アタッチメントの形を変えることで容易に構成を変化させることが可能である.
  Forearmリンクの構造は筋としても骨格としても用いることができるという利点を持った骨構造一体小型筋モジュール(Miniature bone-muscle module)4つによって構成されている.
  筋としてだけでなく骨格になることができる利点を活かし, Bone frameは用いず, Muscle moduleのみで構成された非常に簡易な構成となっている.
  また, 手は切削ばねを用いた柔軟関節構造を有しており\cite{makino2018hand}, 後に詳細を述べる.

  このように, これまで開発したモジュール群とアルミ汎用骨格, そしてそれらを接続するMuscle attachment・Joint attachmentのみにより, 簡易に構成・再構成可能な身体を構築することができる.
}%

\subsubsection{Muscle Configuration}
\begin{figure}[htb]
  \centering
  \includegraphics[width=1.0\columnwidth]{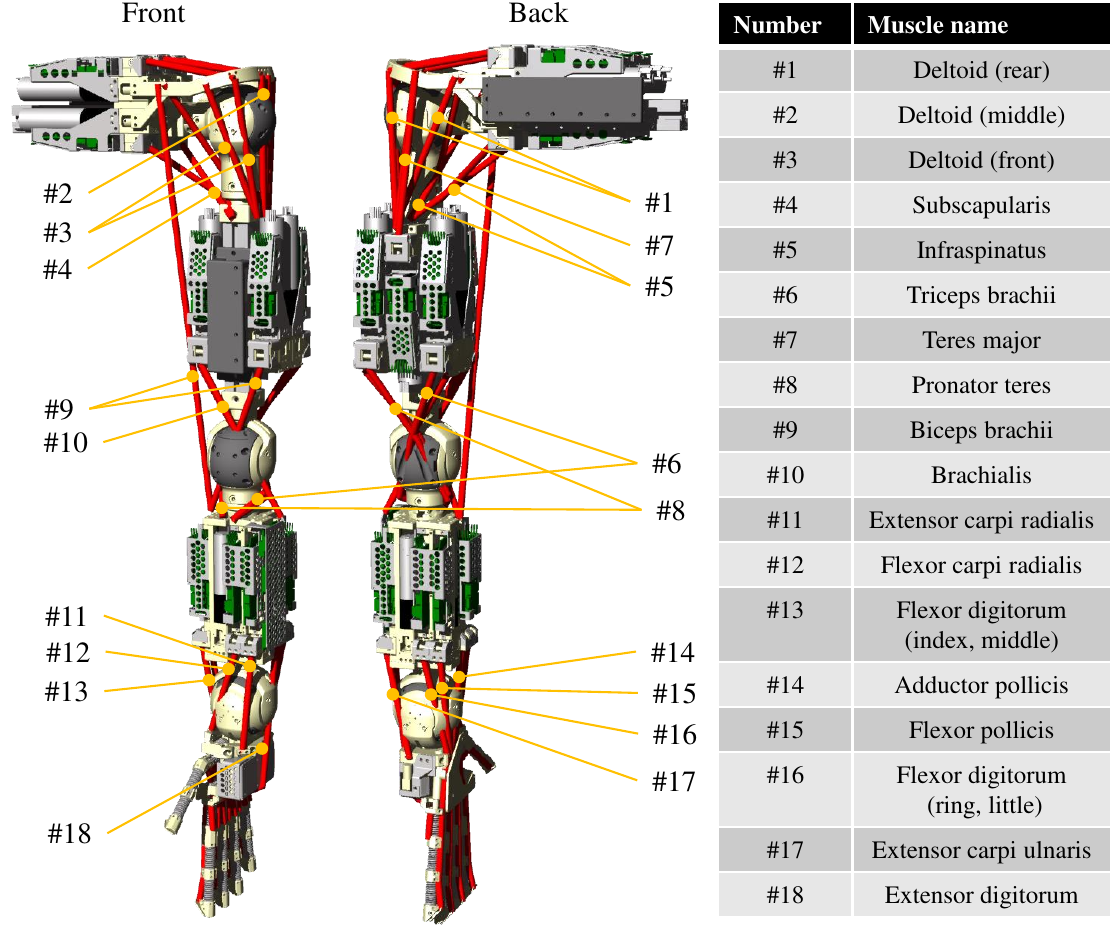}
  \vspace{-3.0ex}
  \caption{Muscle arrangement of MusashiLarm.}
  \label{figure:muscle-arrangement}
  \vspace{-1.0ex}
\end{figure}

\switchlanguage%
{%
  The muscle arrangement of MusashiLarm is shown in \figref{figure:muscle-arrangement}.
  The scapula and humerus include 5 muscles each, the forearm includes 4 miniature bone-muscle modules (8 muscles), and the total number of muscles is 18.
  These muscles imitate the basic muscles of the human body, and include polyarticular muscles.
  The nonlinear elastic units are applied to all muscles except for the ones moving the wrist and fingers, and we use brushless DC motors (90W, gear ratio is 29:1), as actuators.
}%
{%
  MusashiLarmの筋配置を\figref{figure:muscle-arrangement}に示す.
  Scapula・Humerusリンクにそれぞれ5本の筋, Forearmリンクに4つのMiniature bone-muscle module, つまり8本の筋が搭載されており, 合計18本の筋が存在する.
  これらは人体の基本的な筋肉を模倣しており, 多関節筋も有している.
  なお, 本研究で開発された筋構成はForearmリンクについた手首や指を動作させる筋以外に適用されており, アクチュエータとしてはBLDC Motorの90W, ギア比29:1を用いている.
}%

\subsubsection{Circuit Configuration}
\begin{figure}[htb]
  \centering
  \includegraphics[width=1.0\columnwidth]{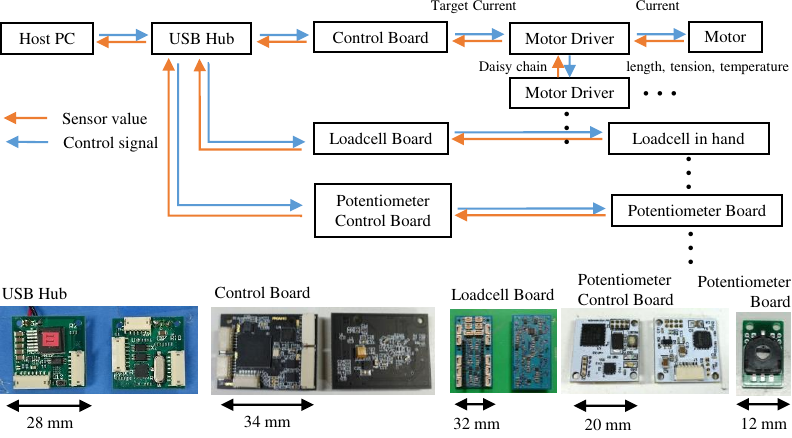}
  \vspace{-3.0ex}
  \caption{Circuit configuration of MusashiLarm.}
  \label{figure:musashilarm-circuit}
  \vspace{-1.0ex}
\end{figure}

\switchlanguage%
{%
  Next, the circuit configuration is shown in \figref{figure:musashilarm-circuit}, and we can connect new modularized components quickly in this configuration.
  The entire communication system is USB communication.
  Control Boards, a Loadcell Board, and Potentiometer Control Boards are connected to USB hubs which are arranged in the body.
  Control Boards are for the muscle motor control, a Loadcell Board amplifies the signal of loadcells attached to the hand and integrates them, and Potentiometer Control Boards integrate the signal of potentiometers included in joint modules.
  Through the Control Board, motor drivers can connect up to 3 drivers by a daisy chain.
  Also, the Loadcell Board can read up to 12 signals of the loadcells, and the Potentiometer Control Board can read up to 4 signals of the potentiometers.
  The Potentiometer Control Board also sends the sensor information of the equipped IMU (MPU9250, InvenSense) to the PC (NUC, Intel), as redundant sensors.
}%
{%
  回路の構成は\figref{figure:musashilarm-circuit}のようになっており, 新しい要素をすぐに繋げられる構成としている.
  全体のシステムはUSB通信によって行われており, 体に配置されたUSBハブにモータ制御のためのControl Board, 手先感覚のためのロードセルの信号を増幅してまとめるLoadcell Board, 関節モジュール内のポテンショメータをまとめるPotentiometer Control Boardが接続している.
  Control Boardは腱悟郎\cite{asano2016kengoro}から変更されており, 全二重通信と半二重通信が両方できるような構成となっている.
  現状Motor Driverが半二重通信のみ可能なため, 通信量の限界からControl Boardを介してMotor Driverは最大3つのみデイジーチェーンで接続可能である.
  今後, より多くデイジーチェーン接続できるよう, Motor Driverを全二重通信が可能なように設計を変更する必要がある.
  また, Loadcell Boardは12個までのロードセルを読むことができ, Potentiometer Control Boardは4つまでのポテンショメータを読むことができる.
  本研究では用いないが, 冗長なセンサ群として, Potentiometer Control Boardはポテンショメータの他に自身に搭載されたIMU(MPU9250, InvenSense, Inc.)の情報も上位に送っている.
}%

\subsection{Musashi: Musculoskeletal Humanoid Platform}
\switchlanguage%
{%
  We can extend MusashiLarm with easily constructable and reconstructable structures to the whole-body musculoskeletal humanoid platform Musashi, by adding only a few joint attachments.
  As another example, this design has been applied to the two-wheel inverted musculoskeletal pendulum, TWIMP \cite{kawaharazuka2018twimp}.
}%
{%
  これまで述べてきたモジュラー化により, 要件である簡易に構成・再構成な身体を構築することができている.
  それにより, このMusashiLarmを全身にまで拡張したMusashiを, Joint attachmentを追加するだけで構成することができる.
}%

\subsubsection{Body Structure}
\begin{figure}[htb]
  \centering
  \includegraphics[width=1.0\columnwidth]{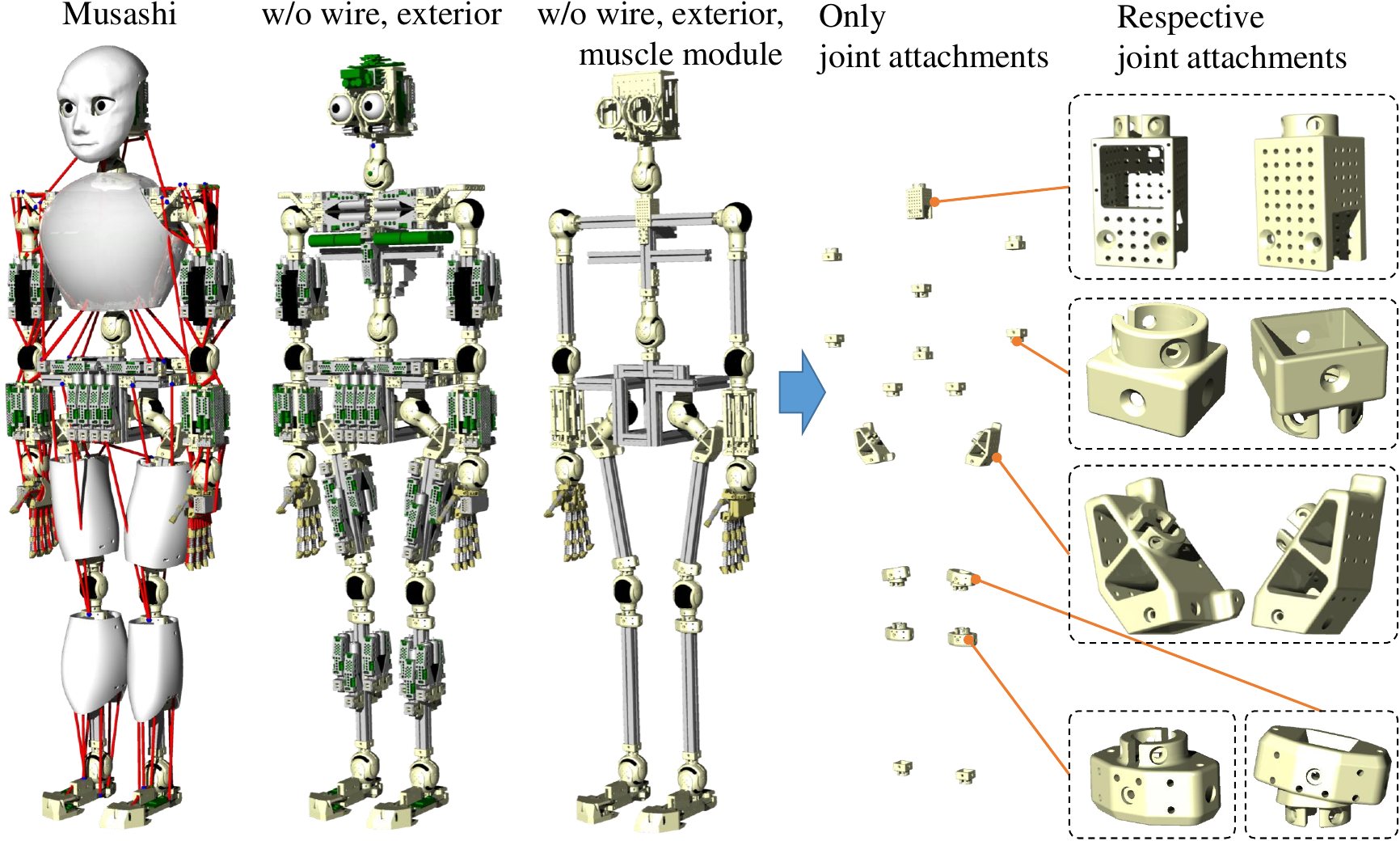}
  \vspace{-3.0ex}
  \caption{Overview of Musashi body structure.}
  \label{figure:musashi-overview}
  \vspace{-1.0ex}
\end{figure}

\switchlanguage%
{%
  We show the musculoskeletal humanoid platform Musashi which extends MusashiLarm, in \figref{figure:musashi-overview}.
  From the left, the figures are the overview, the body without muscle wires, the body without muscle wires and exteriors, and the body without muscle wires, exteriors, and muscle modules.

  We basically extend the structure of MusashiLarm, but the muscle configuration in the lower limb is different from the upper.
  The gear ratio of the muscle actuators in the lower limb is 53:1, and there are no nonlinear elastic units and mechanisms folding back muscle wires.

  Also, we show the body structure of only joint attachments, in the right figure of \figref{figure:musashi-overview}.
  The number of additional joint attachments compared to MusashiLarm is 4, and so we use 5 kinds of joint attachments in Musashi in total.
  Thus, we can construct the whole body of Musashi using joint modules, muscle modules, muscle wire units, generic bone frames, and 6 attachments including the muscle attachment.
  As shown in the left figure of \figref{figure:musashi-overview}, the upper legs, lower legs, chest, and head are covered with exteriors.
}%
{%
  筋骨格上肢MusashiLarmを拡張した筋骨格ヒューマノイドプラットフォームMusashiの全体像を\figref{figure:musashi-overview}に示す.
  左から, 全体像, 筋ワイヤを除いた身体, 筋ワイヤ・外装を除いた身体, 筋ワイヤ・外装・筋モジュールを除いた身体である.

  基本的にはMusashiLarmの構造設計をそのまま全身に拡張しているが, 下半身は筋構成が異なる.
  下半身の筋モジュールのアクチュエータのギア比は全て53:1であり, 非線形弾性要素と折り返し機構が含まれていない.

  また, 筋モジュール・関節モジュール・汎用骨格等を除いたJoint attachmentのみの身体構造を\figref{figure:musashi-overview}の右図に示す.
  MusashiLarmから追加されたJoint attachmentは4種類であり, 上半身と合わせて全5種類を作成している.
  よって, Muscle attachmentを含め, 全6種類のアタッチメントと関節モジュール・筋モジュール・筋ワイヤユニット・アルミ汎用骨格のみで全身の骨格を構成することができた.
  \figref{figure:musashi-overview}に示すように上腿・下腿・胸郭・顔は3Dプリンタで覆われている.
}%

\subsubsection{Head, Hand, and Foot}
\begin{figure}[htb]
  \centering
  \includegraphics[width=1.0\columnwidth]{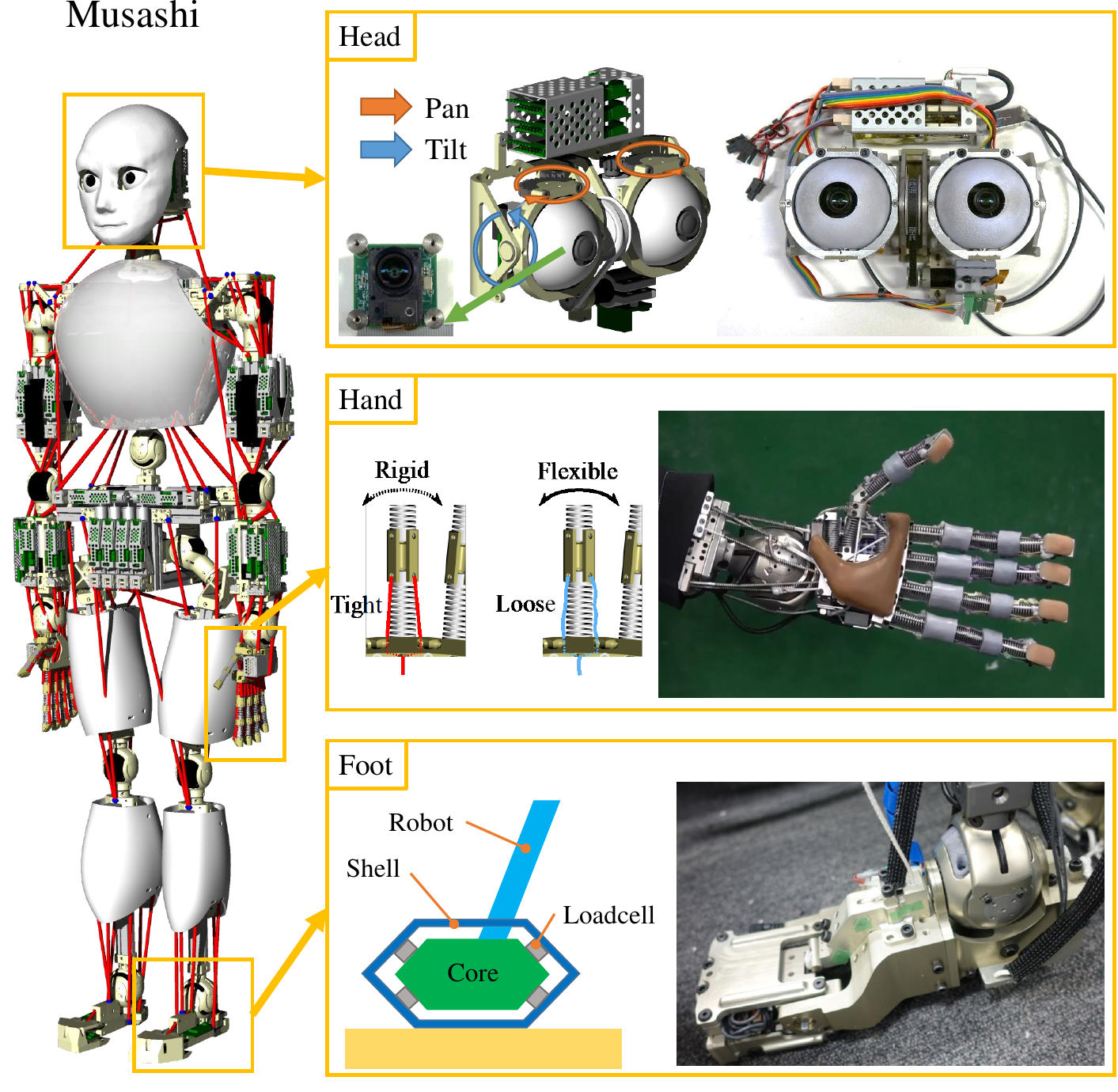}
  \vspace{-3.0ex}
  \caption{Overview of Musashi head, hand, and foot.}
  \label{figure:musashi-characteristics}
  \vspace{-1.0ex}
\end{figure}

\switchlanguage%
{%
  Although we stated the development of Musashi in terms of modularization, in this section, we introduce the characteristic structures of the head, hand, and foot of Musashi (\figref{figure:musashi-characteristics}).

  First, the head equips the movable eye-camera unit \cite{makabe2018eyeunit}, and the camera has various functions such as resolution modification, auto focus, exposure compensation, etc.
  Musashi can recognize the self body by a wide perspective, measure the distance to an object by congestion, and change ROI, etc., and so we can accelerate learning control systems.

  Second, the hand equips flexible and tough five fingers made of machined springs \cite{makino2018hand}.
  The thumb has a wide range of motion by a combination of machined springs, and the fingers have a variable stiffness mechanism.
  Also, the loadcells are equipped in the palm and fingertips, and we can use these sensors to detect environmental contact, as redundant sensors.

  Third, the foot has a core-shell structure, which can measure contact force, in the toe and heel \cite{shinjo2019foot}.
  The structure can measure 6-axis force by the loadcells between the core and shell, and Musashi can measure force not only to the sole but also to the instep.
}%
{%
  モジュール化の観点からMusashiの設計開発について述べたが, 最後にMusashiの頭・手・足の特徴的な機構について説明する(\figref{figure:musashi-characteristics}).

  まず, 頭は可動眼球ユニット\cite{makabe2018eyeunit}を備え, 視覚には解像度や焦点を変更できる多機能なカメラを有している.
  可動眼球の広い視野により自身の身体を認識したり, 輻輳により物体の距離を測定したり, ある注視範囲を高解像度に見たり等により, 学習制御開発を促進する.

  次に, 手は切削ばねによる柔軟で強靭な関節機構を持つ五指ハンド\cite{makino2018hand}となっている.
  親指は切削ばねの組み合わせにより広い可動域を確保し, 指の剛性を変化させる機構を有する.
  また, 手の平と指先にはロードセルを有し, 接触等を検知することで学習制御模索のための冗長なセンサとして使用できる.

  最後に, 足はコアとシェルによって構成された6軸力計測モジュールが踵とつま先についた構造を持つ.
  コア-シェル間の圧力をロードセルによって検出することで6軸力を計測でき, 足裏だけでなく足の甲の力を計測することも可能である.
}%

%%%%%%%%%%%%%%%%%%%%%%%%%%%%%%%%%%%%%%%%%%%%%%%%%%%%%%%%%%%%%%%%%%%%%%%%%%%%%%%%
\section{Experiments} \label{sec:4}
\subsection{Basic Experiments of MusashiLarm}
\begin{figure}[htb]
  \centering
  \includegraphics[width=0.9\columnwidth]{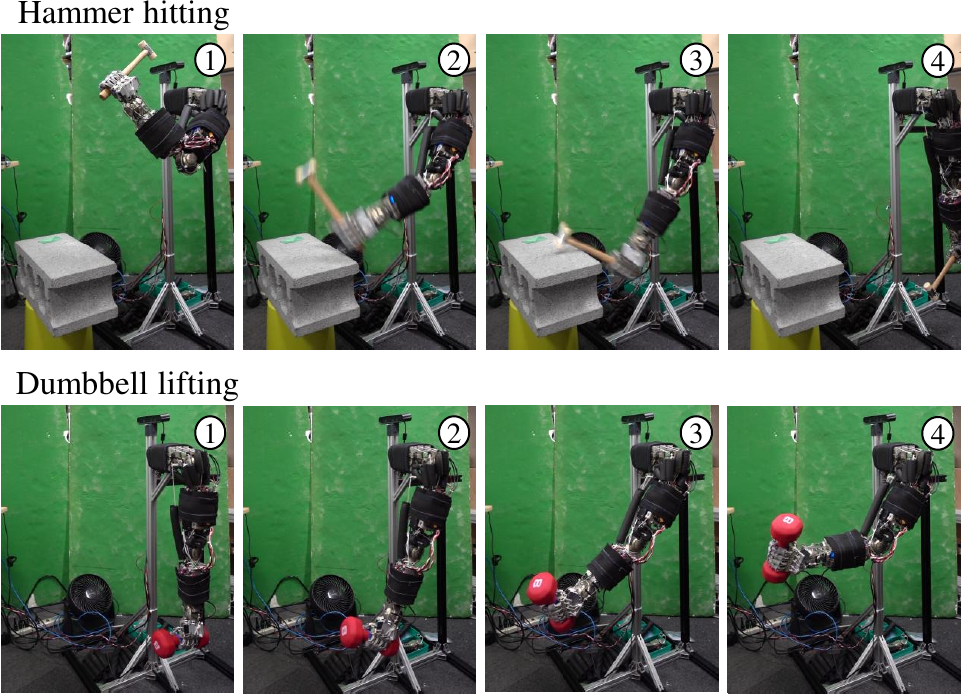}
  \caption{Various motions using MusashiLarm.}
  \label{figure:various-motions}
  \vspace{-1.0ex}
\end{figure}

\switchlanguage%
{%
  First, we conducted a hammer hitting motion with a block.
  The result is shown in the center figure of \figref{figure:various-motions}.
  Due to the soft body, the upper limb can absorb external impact and parry force without breaking the robot itself.

  Second, we conducted a dumbbell lifting motion with a dumbbell weight of 3.6 kg.
  The result is shown in the upper figure of \figref{figure:various-motions}.
  Although the upper limb has a very soft body, it can exhibit high power and succeeded in lifting the dumbbell.
}%
{%
  MusashiLarmを用いていくつかの基礎実験を行う.

  まず、ハンマーでブロックを叩く動作を行った。
  その様子を\figref{figure:various-motions}の中図に示す。
  柔らかい体を持つことで、衝撃を吸収し、ハードウェアを故障させることなく力を受け流すことができている。

  次に、約3.6kgのダンベルを持ち上げる動作を行った。
  その様子を\figref{figure:various-motions}の上図に示す。
  3.6kgのダンベルを持ち上げることに成功し、弾性要素によって柔らかい体を持ちながらも、強い力を出すことができている。
}%

\subsection{Learning Control of Handle Operation}
\begin{figure}[htb]
  \centering
  \includegraphics[width=1.0\columnwidth]{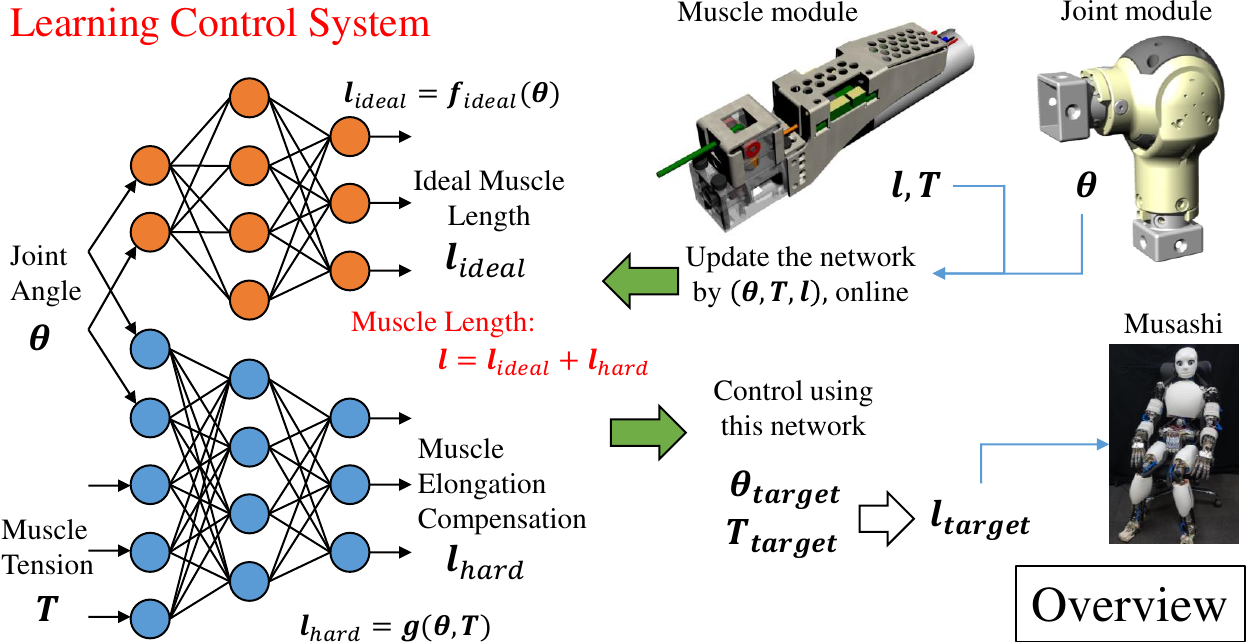}
  \vspace{-3.0ex}
  \caption{The overview of the learning control system extending the previous method \cite{kawaharazuka2018bodyimage}.}
  \label{figure:learning-control}
  \vspace{-1.0ex}
\end{figure}

\begin{figure}[htb]
  \centering
  \includegraphics[width=1.0\columnwidth]{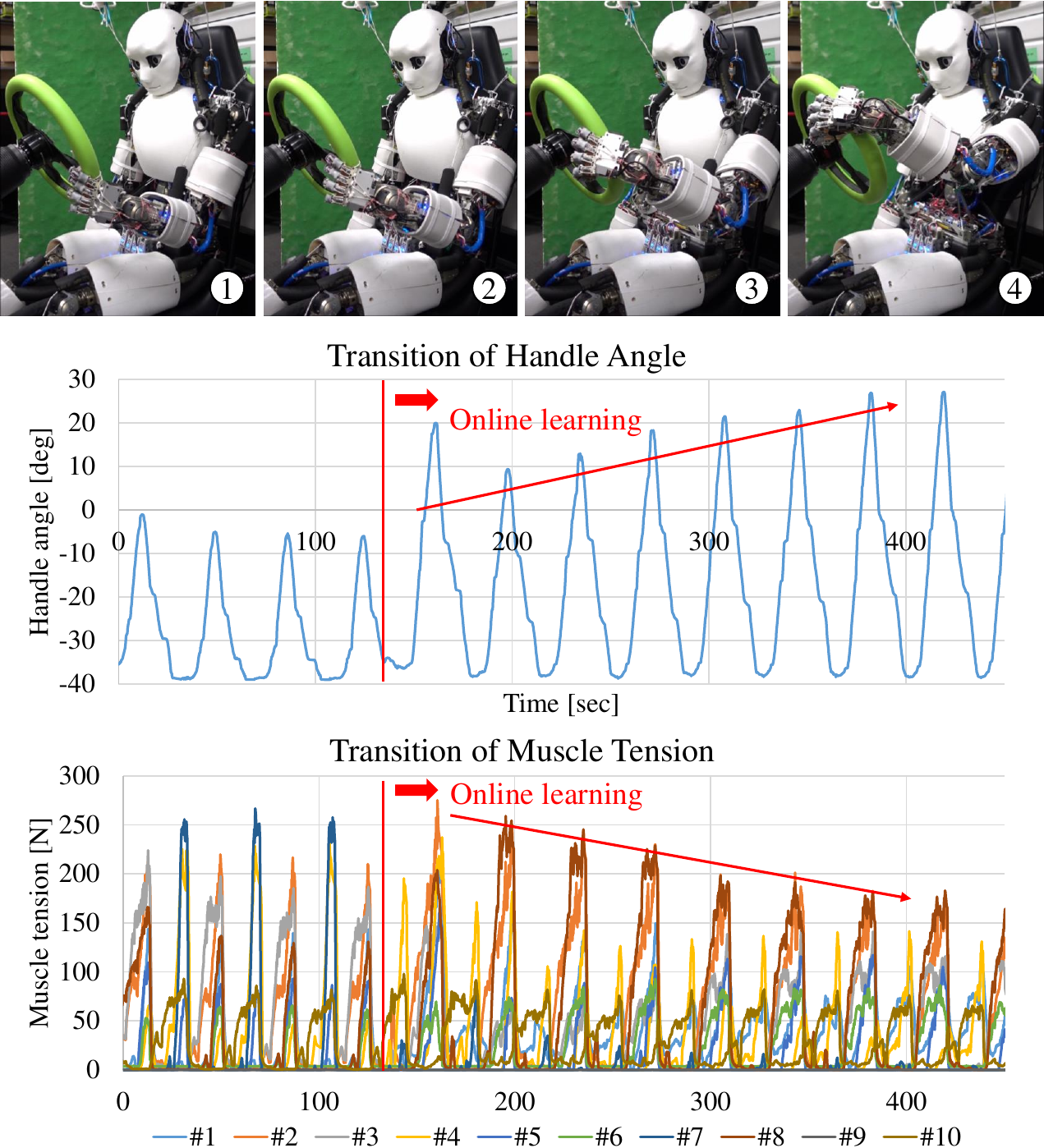}
  \vspace{-3.0ex}
  \caption{The experiment of the handle operation. Lower graphs express the transition of the handle rotation angle and muscle tensions.}
  \label{figure:handle-operation}
  \vspace{-1.0ex}
\end{figure}

\switchlanguage%
{%
  We conducted a handle operation experiment with Musashi, as an example of the environmental contact behavior by a flexible body using a learning control system.
  In the handle operation, the arm must move along the circular orbit and needs a large force.
  Also, we must accurately describe the relationship between the handle rotation and movement of the flexible arm.

  Therefore, we use a learning control method \cite{kawaharazuka2019longtime} (\figref{figure:learning-control}) which improved on the self-body image acquisition \cite{kawaharazuka2018bodyimage}, and the robot acquires handle operation while moving.
  We update the self-body image using the joint angles measured from the joint modules, and the muscle lengths and tensions measured from muscle modules.
  Then, we can conduct position control by calculating the target muscle length by inputting the target joint angles and measured muscle tensions into the self-body image.
  Musashi rotates the handle from -30 deg to 60 deg while solving inverse kinematics, and the self-body image is updated while moving.
  We can measure the rotation of the handle from AR marker.
  The experimental appearance is shown in the upper figure of \figref{figure:handle-operation}, and the transition of the handle rotation and muscle tensions is shown in the lower figure.
  In the beginning, the muscle tension is high at 270 N, and the handle rotates only about 40 deg in total.
  By updating the self-body image online, the relationship between joint angles, muscle lengths, and muscle tensions when rotating the handle, is learned.
  Finally, the handle rotates about 65 deg in total, and the maximum muscle tension to rotate it is about 180 N, because the difference between the actual robot and its geometric model including the antagonistic relationship is correctly modified.
  We were able to conduct an example of learning control systems using the redundant sensors and flexible structure of Musashi.
  However, it is difficult to completely conduct intended movements, and we would like to investigate learning controls considering hysteresis, etc., in the future.
}%
{%
  学習制御を用いた柔軟な身体による環境接触行動の例として, Musashiによる自動車のハンドル操作を行う.
  ハンドル操作は, 身体を座席に固定した状態で, ハンドルを回すような円軌道をマニピュレータに描かせる必要がある.
  環境接触を伴い, 特にハンドルを回転させるためには力が必要であり, 柔軟な身体における力のかけ具合とハンドルの回転を正確に記述しなければならない.
  そこで本研究では, 自己身体像の逐次獲得手法\cite{kawaharazuka2018bodyimage}を改良した手法(\figref{figure:learning-control}) \cite{kawaharazuka2019longtime}により, 動作させながら, ハンドル操作を獲得していく.
  本手法では, 冗長なセンサとしての関節モジュールから得られる関節角度と筋モジュールから得られる筋張力・筋長のデータにより, 初期学習された自己身体像をオンラインで更新していく.
  また, 自己身体像に目標関節角度と現在筋張力を入力し, 指令筋長を得るような位置制御を行う.
  Musashiの左手のみを用いて, ハンドルを-30 degから60 degまで30 degずつ回転させる動作を逆運動学を解きながら連続して行い, その際に常に自己身体像を更新し続ける.
  ハンドルにはAR markerをつけ, 常にその角度を測定する.
  実験の際の様子を\figref{figure:handle-operation}の上図に, 実験中のハンドル角度, 筋張力の推移を\figref{figure:handle-operation}の下図に示す.
  はじめは動作の際の筋張力が270 N程度と高く, ハンドルの角度も全体で40 deg程度しか回転していない.
  自己身体像獲得により, 環境であるハンドルとの接触からその際にかかる筋張力と筋長・関節角度の関係が学習されていき, 最終的には筋張力は180 N程度, ハンドルは全体で65 deg程度回転するようになる.
  しかし, 学習後も完全に意図した動作は難しく, 今後ヒステリシス等を考慮した学習制御の模索を行っていきたい.

}%

%%%%%%%%%%%%%%%%%%%%%%%%%%%%%%%%%%%%%%%%%%%%%%%%%%%%%%%%%%%%%%%%%%%%%%%%%%%%%%%%
\section{CONCLUSION} \label{sec:5}
\switchlanguage%
{%
  In this study, we stated the component modularized design of the musculoskeletal humanoid platform Musashi, to investigate learning control systems.
  At first, we set the flexible body, redundant sensors, and easily reconfigurable structure, as the requirements, and designed respective modules to fulfill them.
  We proposed generic joint modules including potentiometers and IMU, which can be applied to various joints and be used for redundant sensors in learning control systems.
  We designed 2 kinds of compact and reliable muscle modules, and used them depending on different body parts.
  Regarding muscle wire units, we standardized muscle relay units by the direction of the folded back muscle and the mounting direction, and realized compact and strong nonlinear elastic units with a soft structure, by using the compression of the grommet structure.
  Then, we designed MusashiLarm composed of only joint modules, muscle modules, muscle wire units, generic bone frames, and a few attachments, and designed a musculoskeletal humanoid platform Musashi by extending MusashiLarm to the whole body with only 4 additional joint attachments.
  Also, we conducted experiments with environmental contact and using learning control systems, and verified the effectiveness of our concepts.
  In future works, we would like to investigate more advanced learning control systems using these musculoskeletal platforms.
}%
{%
  本研究では, 学習制御を模索するための筋骨格ヒューマノイドプラットフォームMusashiにおけるコンポーネントのモジュール化について述べた.
  その要件として, 柔軟な身体・冗長なセンサ・簡易な組み換えを挙げ, それらを可能とするようなモジュール設計を行った.
  関節モジュールはポテンショメータ・IMUを内包したコンパクトな設計を提案し, 学習制御のための冗長なセンサとして使用できる.
  筋モジュールは二種類設計し, 部位によって使い分けることでコンパクトで信頼性のある設計をすることができる.
  筋ワイヤユニットは, 折り返し機構を筋の方向と装着方向の組み合わせから標準化し, 非線形弾性ユニットはグロメット構造を圧縮することでそれ自体が柔らかく信頼性のある形で実現した.
  これらの関節モジュール・筋モジュール・筋ワイヤユニット・汎用骨格・少数のアタッチメントのみで構成可能なMusashiLarmを設計し, これを全身に拡張した筋骨格ヒューマノイドプラットフォームMusashiを4つのJoint attachmentの追加のみで構成することに成功した.
  また, 柔軟な身体を用いた環境接触行動と学習制御によるハンドル操作を行い, 本研究のコンセプトの有効性を確認した.
  今後はこれらを用いた高度な学習制御の模索に力を入れたい.
}%

{
  %\footnotesize
  %\small
  %\bibliographystyle{junsrt}
  \bibliographystyle{IEEEtran}
  \bibliography{bib}
}

\end{document}